\documentclass[preprint,sort,compress,12pt]{elsarticle}
\usepackage[T1]{fontenc}
\usepackage[utf8]{inputenc}
\usepackage{graphicx}
\usepackage{amsmath,amssymb,bm,mathrsfs}
\usepackage{array,upgreek,float,booktabs,adjustbox,makecell}
\usepackage[margin=1in]{geometry}
\usepackage[small,compact]{titlesec}
\usepackage[normal,bf,labelsep=colon,skip=2pt]{caption}
\usepackage[titles]{tocloft}
\usepackage{bold-extra,multirow,lineno,tabularx,diagbox}
\usepackage{algorithm,algorithmic}
\usepackage{subcaption}
\usepackage{microtype}  

\usepackage{xcolor}
\usepackage{hyperref}
\biboptions{sort&compress}
\hypersetup{
    colorlinks=true,
    linkcolor={red!50!black},
    citecolor={blue!50!black},
    urlcolor={blue!80!black},
    pdftitle={Score-Based Stochastic Closure Modeling via Indirect Learning and Online Data Assimilation}
}
\AtBeginDocument{\setlength{\abovedisplayskip}{8pt plus 2pt minus 2pt}\setlength{\belowdisplayskip}{8pt plus 2pt minus 2pt}}
\begin{document}
\begin{frontmatter}
\title{Score-Based Stochastic Closure Modeling via Offline Indirect Learning and Online Data Assimilation}

\author{Xinghao Dong}
\author{Huchen Yang}
\author{Jin-Long Wu\corref{cor1}} \ead{jinlong.wu@wisc.edu} 
\cortext[cor1]{Corresponding author}

\address{Department of Mechanical Engineering, University of Wisconsin–Madison, Madison, WI 53706}

\begin{abstract}
Stochastic closure modeling is an emerging area in reduced-order modeling of large and complex (e.g., multiscale, multi-physics) systems, for which the computationally affordable reduced-order physical descriptions (e.g., digital twins for additive manufacturing, world models for AI agents) in numerical simulations inevitably lead to information losses that cannot be well addressed by classical deterministic closures. However, learning stochastic closures typically requires closure data that are difficult to obtain from measurements or often prohibitively expensive to compute using first-principles simulations. In this work, we develop a stochastic closure modeling framework that builds on score-based diffusion models, leveraging indirect offline training and online data assimilation without requiring the direct data of closures. Numerical experiments on two-dimensional turbulent flows demonstrate that indirect offline learning captures key closure statistics and improves predictions of system state. In addition, combining the learned closure with known physical dynamics leads to better generalization capability across different regimes than purely data-driven models that directly predict system state. Compared to standard data assimilation methods with a Gaussian assumption, the posterior-score assimilation better represents non-Gaussian features of the learned closure, and its combination with state-level data assimilation further improves prediction of the system state. The proposed framework enables stochastic closure modeling in realistic and challenging scenarios of complex dynamical systems in mechanics and engineering problems, where non-Gaussian statistics of model-form uncertainties are non-negligible (e.g., due to the nonlinearities of the system dynamics and/or the observations) and true closure data are not available, for both offline learning and online assimilation.
\end{abstract}

\end{frontmatter}

\section{Introduction}\label{sec:introduction}

Applications ranging from weather and climate prediction to molecular simulation and robotic control require models that predict complex dynamics at an affordable computational cost \citep{kochkov2024neural,charron2025navigating,wu2023daydreamer}. For multiscale physical systems, meeting this requirement often calls for reduced descriptions that retain selected scales or variables \citep{majda2014new, peherstorfer2016operator}, with closure models accounting for the influence of unresolved interactions on the resolved dynamics \citep{pope2000turbulent,sagaut2006large,meneveau2000scale,moser2021statistical}. Classical closures typically parameterize these interactions through physical reasoning and empirical calibration, while recent machine-learning approaches can learn the dependence of closure terms on resolved quantities from data \citep{smagorinsky1963general,germano1991dynamic,lilly1992proposed,duraisamy2019turbulence,ling2016reynolds,rasp2018deep,zhou2021nonlocal,maulik2019subgrid,beck2019deep}. Many of these models assign a single deterministic closure to each resolved state, although different unresolved fine-scale fields can share the same resolved state and have different effects on its subsequent evolution, making the uncertainties due to the unresolved fine-scale fields non-negligible~\citep{palmer2019stochastic}. The information lost through unresolved scales can therefore introduce model-form uncertainty into the resolved physical description \citep{mori1965transport,zwanzig1961memory,chorin2000optimal,chorin2002optimal,wouters2013multi}, motivating stochastic closures that represent unresolved fluctuations alongside their average influence \citep{buizza1999stochastic,palmer2001nonlinear,wilks2005effects,franzke2015stochastic,berner2017stochastic,kondrashov2015data}.

Machine learning can support prediction of these systems either by learning the resolved evolution directly or by learning closure terms within a reduced-order physical model. Purely data-driven models approximate state transitions or solution trajectories without explicitly solving the governing equations during prediction, making them useful when governing equations are unavailable or too costly to solve numerically \citep{lu2021learning,kovachki2023neural,chen2025neuraldynamical}. Hybrid models combine known governing equations with learned components, including closure terms that represent unresolved interactions \citep{menier2023cdrom,pathak2018hybrid,brunton2020machine,duraisamy2019turbulence}. Learning these closure terms while retaining the known dynamics can reduce what must be inferred from training data and support generalization beyond the training regime \citep{barsinai2019learning,kochkov2021machine}.

Within this hybrid approach, stochastic closure learning targets a conditional distribution of closure realizations given the resolved state, its history, or other available information \citep{crommelin2008subgrid,kondrashov2015data}. Learning such distributions and generating samples from them are central tasks in generative artificial intelligence (GenAI). Diffusion models, including score-based formulations, have been extensively developed for image synthesis and restoration \citep{sohl2015deep,song2019generative,ho2020denoising,song2021score,karras2022elucidating}. These methods have also been applied to probabilistic forecasting and reconstruction of physical fields \citep{cachay2023dyffusion,kohl2026benchmarking,price2025probabilistic,gao2025generative,fan2026generative,turan2026toward}. GenCast, for example, uses conditional diffusion to generate ensembles of global weather trajectories from current and previous atmospheric states \citep{price2025probabilistic}. Within closure modeling, generative adversarial networks and variational autoencoders have been used to learn unresolved forcing and spatially correlated subgrid effects \citep{gagne2020machine,perezhogin2023generative}. \citet{dong2025data} combined conditional diffusion with neural operators to construct stochastic, spatially nonlocal closure models. Subsequent latent-space and transport-based formulations improved sampling efficiency and examined how learned representations affect closure fidelity \citep{dong2026latent,dong2026transport}.

Directly supervised training of stochastic closure models requires reference closure fields. For example, \citet{dong2025data} trained conditional diffusion models using state fields paired with reference closures computed from numerical simulations. Such closure data are often difficult to obtain from measurements or expensive to compute through sufficiently detailed simulations \citep{perezhogin2023generative,pope2000turbulent,sagaut2006large}. When only resolved-state snapshots are available, estimating closures from the governing equations requires numerical differentiation in time, which can be unreliable for sparsely sampled data \citep{wu2024learning}. These limitations motivate indirect learning of stochastic closures through their effects on the resolved dynamics, drawing on inverse modeling and model calibration \citep{stuart2010inverse,parish2016paradigm,iglesias2013ensemble,cleary2021calibrate,wu2024learning,yang2026bayesian,lourenco2024indirect}. For stochastic modeling, \citet{schneider2021learning} used ensemble Kalman inversion to infer drift and diffusion functions by matching simulated statistics to statistics derived from observed time series. \citet{wu2024learning} developed a broader formulation for learning structural-error models embedded within governing dynamics, including deterministic, stochastic, and nonlocal corrections. Gradient-based learning through numerical integration provide a complementary route \citep{chen2018neural,kidger2021neural,rasp2020coupled,kochkov2021machine}. In particular, \citet{um2020solver} trained neural correction models within differentiable PDE solvers, accounting for how each correction changes subsequent states and the inputs encountered by later predictions.

Beyond offline learning, data assimilation (DA) can improve reduced-order predictions by incorporating observations as the system evolves \citep{zerfas2019continuous,plogmann2026sparse}. For probabilistic models, the learned distribution can usually serve as a prior for Bayesian DA, which combines this prior with the observation likelihood to infer a posterior distribution \citep{reich2015probabilistic,carrassi2018data,chen2025discretecgkn}. In score-based generative models, posterior sampling can be performed by adding likelihood guidance to the learned prior score during reverse diffusion, directing samples toward consistency with both the learned statistics and the measurements \citep{song2021score,chung2023diffusion,song2023pseudoinverse}. These methods have been applied to noisy inverse problems \citep{dasgupta2025conditional,song2023pseudoinverse,kawar2022denoising} and data assimilation in dynamical systems \citep{bao2024score,bao2024ensemble,manshausen2025generative}. For example, \citet{chung2023diffusion} used denoised estimates to approximate likelihood guidance for noisy linear and nonlinear inverse problems, while \citet{rozet2023score} learned trajectory priors that support assimilation of incomplete observations without retraining the generative model for each observation setting. 

Although score-based DA can generate individual posterior samples without ensemble covariance estimates, sample-based uncertainty quantification generally calls for multiple simulated trajectories \citep{reich2015probabilistic,owen2013monte,talagrand1997evaluation}, providing ensembles that can also support ensemble Kalman filtering (EnKF). The EnKF uses sample cross-covariances between the variable being updated and predicted observations, allowing observations to inform quantities that are not directly measured \citep{evensen1994sequential,evensen2003ensemble,burgers1998analysis}. Its covariance-based analysis is generally well suited to approximately linear--Gaussian conditions, although nonlinear observations and non-Gaussian priors can produce posterior structure that these updates do not fully capture \citep{anderson2010non,lei2011moment}. Previous studies have used diffusion models to generate ensembles for subsequent Kalman analysis, connecting generative sampling with established ensemble-based assimilation methods \citep{asahi2023generating,finn2024representation}.

Building on these developments in indirect learning and data assimilation, we develop a conditional score-based framework for stochastic closure modeling from resolved-state data only. The framework learns closures offline from recorded state trajectories and refines predictions online as new observations become available. The main contributions are summarized as follows:
\begin{itemize}
    \item We formulate indirect training of conditional stochastic closures by a differentiable trajectory energy score loss through the reverse-diffusion sampler and reduced-order physical solver. The training objective accounts for agreement with the resolved-state data and variability among generated trajectories, enabling learning without direct data of reference closure fields or estimated time derivatives.

    \item We evaluate closure fidelity and predictions of resolved states through a priori analyses and a posteriori simulations of two-dimensional turbulence, using directly supervised closures as a baseline for indirect learning. Comparisons with purely data-driven models at comparable short-window accuracy in prediction show the benefits of retaining resolved physics for generalization across the tested physical regimes.

    \item We develop posterior-score DA for closures at the model-form uncertainty level, with comparisons against closure EnKF showing improved representation of non-Gaussian statistics under nonlinear observations and better agreement in closure statistics during sequential assimilation with linear state observations. Combining this closure analysis with state EnKF via a progressive correction approach~\cite{musso2001improving,frei2013} further improves prediction accuracy and uncertainty calibration.
\end{itemize}

Section~\ref{sec:methodology} presents the formulation and algorithms, Section~\ref{sec:results} evaluates closure fidelity, regime generalization, and online assimilation, and Section~\ref{sec:conclusions} summarizes the findings.

\section{Methodology}\label{sec:methodology}

Consider a multiscale dynamical system with state $v(\mathbf{x},t)$ on a spatial domain $\Omega$, and let $\mathcal{M}$ denote its governing operator. A fixed linear coarse-graining operator $\mathcal{P}$ defines the resolved state $V=\mathcal{P}v$. Since coarse-graining generally does not commute with nonlinear operators, evaluating the governing equations from the resolved state alone does not reproduce their filtered fine-scale effects \citep{pope2000turbulent,sagaut2006large}. Let $\mathcal{R}[V]$ denote the dynamics described by the reduced-order physical model. We define the closure field as the difference between $\mathcal{R}[V]$ and the coarse-grained full-system dynamics:
\begin{equation}\label{eq:general_closure_definition}
    \Pi := \mathcal{R}[\mathcal{P}v] - \mathcal{P}\mathcal{M}[v].
\end{equation}
The resolved evolution then takes the form
\begin{equation}\label{eq:general_resolved_dynamics}
    \frac{\partial V}{\partial t} = \mathcal{R}[V]-\Pi.
\end{equation}
While $\mathcal{R}$ can be evaluated from the resolved state, evaluating $\Pi$ generally requires fine-scale information not retained in the reduced-order model. The two-dimensional turbulence experiments in Section~\ref{ssec:numerical_setup} use spatially filtered vorticity as the resolved state, with the filtering procedure and corresponding closure specified in Section~\ref{sssec:filters}.

We consider two training settings for stochastic closure models, distinguished by the available data. Direct training uses paired resolved fields and reference closures, whereas indirect training uses only recorded resolved trajectories:
\begin{align}
    \mathcal{D}_{\mathrm{pair}} &= \left\{ \left(V_k^{\mathrm{ref}},\Pi_k^{\mathrm{ref}}\right) \right\}_{k=1}^{N_{\mathrm{pair}}}, \label{eq:direct_training_data}\\
    \mathcal{D}_{\mathrm{state}} &= \left\{ \left( V_i^{\mathrm{ref}}(t_0),\ldots, V_i^{\mathrm{ref}}(t_J) \right) \right\}_{i=1}^{N_{\mathrm{tr}}}. \label{eq:resolved_training_data}
\end{align}
Here, the superscript $\mathrm{ref}$ denotes reference data, and $N_{\mathrm{pair}}$ and $N_{\mathrm{tr}}$ denote the numbers of paired samples and training trajectories, respectively. The snapshot times satisfy $t_0<\cdots<t_J$. In the indirect setting, each snapshot contains the full resolved field, but the records may be sparse in time, with many numerical integration steps between successive snapshots. The recorded trajectories provide indirect information about closures through the accumulated effects of unresolved interactions on the resolved evolution, enabling training without reference closure fields or estimated time derivatives of the data.

During prediction, new measurements provide further information about the evolving system. We represent them as $y_j=\mathcal{H}_j[V^{\mathrm{ref}}(t_j)]+\epsilon_j$, where $\mathcal{H}_j$ maps the resolved state to the measured quantity and $\epsilon_j$ represents observation error. These measurements may be partial and noisy. Through the resolved dynamics, they can refine the sampled closure distribution indirectly at the model-form uncertainty level and can also correct the predicted state directly at the prediction level. The formulation thus connects offline closure learning with online assimilation through the same resolved-system evolution.

\subsection{Hybrid modeling with model-form uncertainty}
\label{ssec:hybrid_formulation}

We model the closure effects missing from the reduced-order physical description through a conditional distribution, providing a stochastic representation of model-form uncertainty \citep{xiao2016quantifying,wu2024learning,zhang2025probabilistic}:
\begin{equation}\label{eq:conditional_closure_model}
    \Pi_\theta\sim p_\theta(\Pi\mid V),
\end{equation}
where $\theta$ denotes the trainable model parameters and $V$ supplies the resolved-state information available for each closure draw. For a given resolved state, the conditional mean describes the average closure field, while the spread, spatial correlations, and higher-order statistics describe the variability of closure realizations \citep{dong2025data}. Learning this distribution accounts for both the average correction to the resolved dynamics and the fluctuations associated with unresolved interactions.

The hybrid model couples the learned stochastic closure with the retained physical dynamics by replacing $\Pi$ in Eq.~\eqref{eq:general_resolved_dynamics} with a sampled closure $\Pi_\theta$:
\begin{equation}\label{eq:hybrid_resolved}
    \frac{\partial V_\theta}{\partial t} = \mathcal{R}[V_\theta]-\Pi_\theta,
\end{equation}
where $V_\theta$ denotes the trajectory produced by the closure model and physical solver. During prediction, the trained parameters remain fixed, while each new closure draw uses the current resolved state along the corresponding trajectory. Different closure realizations can therefore produce different trajectories from a common initial field, and the resulting state differences influence subsequent closure sampling. An ensemble of these coupled simulations describes how closure variability propagates to prediction uncertainty. 

Purely data-driven prediction models instead learn a state-transition map that approximates the resolved evolution in Eq.~\eqref{eq:general_resolved_dynamics} over a prediction interval $\Delta t$ \citep{brunton2020machine}:
\begin{equation}\label{eq:data_driven_transition}
    V_\vartheta(t+\Delta t) = \mathcal{F}_\vartheta \left(V_\vartheta(t),\xi\right),
\end{equation}
where $\vartheta$ denotes the trainable parameters of the transition map and $\xi$ is a random input for probabilistic prediction. Omitting $\xi$ gives a deterministic predictor, while repeated sampling generates an ensemble of future states from the same current resolved state. The learned map represents the complete state transition, including predictive variability in the probabilistic formulation. In the hybrid model of Eq.~\eqref{eq:hybrid_resolved}, learning instead targets the closure distribution, with the physical solver propagating closure variability into resolved-state predictions \citep{dong2025data}.

\begin{figure}[!htb]
    \centering
    \includegraphics[width=\linewidth]{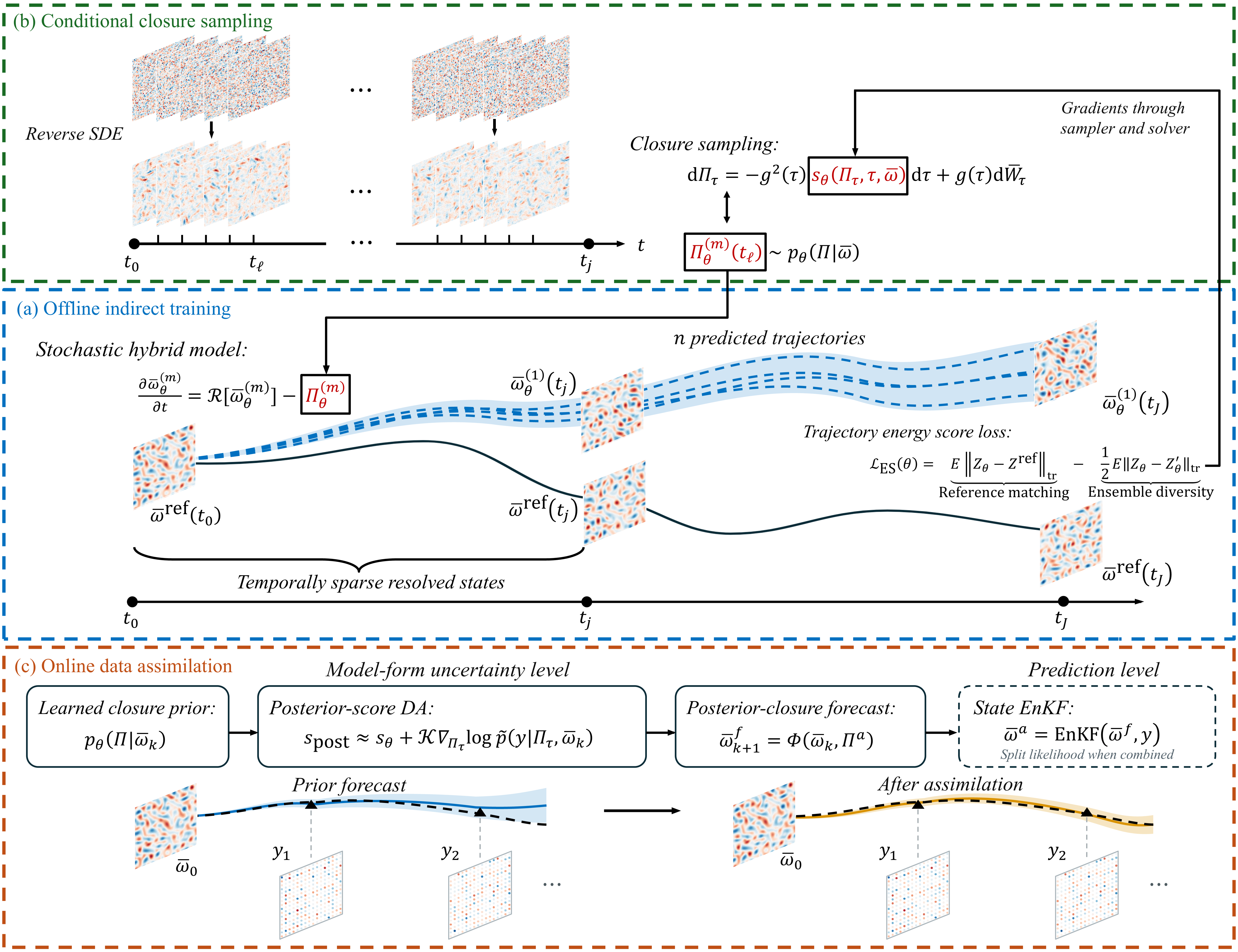}
    \caption{Schematic of the proposed framework. (a) Offline indirect training compares simulated resolved trajectories with temporally sparse reference states through a trajectory energy score loss. (b) Reverse diffusion generates closure fields conditioned on the evolving resolved state. (c) Online posterior-score DA refines closures at the model-form uncertainty level, while complementary state EnKF corrects the resolved prediction. Black curves illustrate reference trajectories, and colored curves and shaded bands represent ensemble predictions and their uncertainty.}
    \label{fig:framework_schematic}
\end{figure}

Both formulations can incorporate prescribed physical parameters and external inputs. Purely data-driven models learn how these inputs affect the state transition, while the hybrid solver evaluates known physical contributions through $\mathcal{R}$ and uses the learned closure to represent effects missing from the reduced-order description. We learn and sample the conditional closure distribution $p_\theta(\Pi\mid V)$ using a score-based generative model, which supplies the stochastic component of the framework illustrated in Fig.~\ref{fig:framework_schematic}.

\subsection{Score-based stochastic closure modeling}
\label{ssec:conditional_score}

Following the conditional score-based closure formulation of \citet{dong2025data}, we learn a conditional score for generating closure fields from Gaussian noise. We distinguish physical time $t$ from diffusion time $\tau\in[0,1]$. For paired resolved states and clean closures $(V,\Pi_0)$, forward diffusion perturbs the closure while keeping $V$ fixed:
\begin{equation}\label{eq:ve_forward}
    \mathrm{d}\Pi_\tau = g(\tau)\,\mathrm{d}W_\tau, \qquad \sigma^2(\tau) = \int_0^\tau g(s)^2\,\mathrm{d}s,
\end{equation}
where $W_\tau$ is a standard Wiener process and $g$ determines the noise schedule. Equivalently, $\Pi_\tau=\Pi_0+\sigma(\tau)\xi$, with$\xi\sim\mathcal{N}(0,I)$. Averaging this Gaussian perturbation over the clean closure distribution at fixed $V$ gives the noisy conditional density $p_\tau(\Pi_\tau\mid V)$. Its score is approximated by a network taking the noisy closure, diffusion time, and resolved state as inputs:
\begin{equation}\label{eq:conditional_closure_score}
    s_\theta(\Pi_\tau,\tau,V) \approx \nabla_{\Pi_\tau}\log p_\tau(\Pi_\tau\mid V).
\end{equation}

Direct training learns this conditional score from jointly sampled state--closure pairs through denoising score matching \citep{vincent2011connection,dong2025data}:
\begin{equation}\label{eq:direct_dsm}
    \mathcal{L}_{\mathrm{DSM}}(\theta) = \mathbb{E}_{(V,\Pi_0),\tau,\xi} \left[ \left\| \sigma(\tau) s_\theta\!\left( \Pi_0+\sigma(\tau)\xi,\tau,V \right) +\xi \right\|_2^2 \right],
\end{equation}
where $(V,\Pi_0)$ is drawn from $\mathcal{D}_{\mathrm{pair}}$ and $\tau$ is sampled over the prescribed positive noise range. The Gaussian perturbation provides an analytic denoising target, allowing the conditional score to be learned without evaluating the conditional density. This formulation nevertheless requires reference closure fields to construct both the noisy inputs and training targets.

Closure sampling reverses the forward diffusion while keeping the resolved state $V$ fixed. At sufficiently large noise, the noisy closure distribution is approximated by an isotropic Gaussian, providing the initialization $\Pi_1=\sigma(1)\xi_{\mathrm{init}}$, with $\xi_{\mathrm{init}}\sim\mathcal{N}(0,I)$. The learned score then guides the reverse-time SDE \citep{anderson1982reverse,song2021score,dasgupta2026unifying}:
\begin{equation}\label{eq:ve_reverse}
    \mathrm{d}\Pi_\tau = -g(\tau)^2 s_\theta(\Pi_\tau,\tau,V)\,\mathrm{d}\tau + g(\tau)\,\mathrm{d}\overline{W}_\tau, \qquad \tau:1\longrightarrow0,
\end{equation}
where $\overline{W}_\tau$ is a Wiener process in reverse diffusion time.

We discretize this process over decreasing diffusion times $\tau_0=1>\cdots>\tau_K=\varepsilon$, where $\varepsilon>0$ is a small terminal noise cutoff. Defining $\Delta\sigma_k^2 =\sigma^2(\tau_k)-\sigma^2(\tau_{k+1})>0$, the Euler--Maruyama update in variance time is
\begin{equation}\label{eq:reverse_variance_step}
    \Pi_{\tau_{k+1}} = \Pi_{\tau_k} + \Delta\sigma_k^2 s_\theta(\Pi_{\tau_k},\tau_k,V) + \sqrt{\Delta\sigma_k^2}\,\xi_k,
\end{equation}
where $\xi_k$ are standard Gaussian random fields, independent across reverse-diffusion steps. The resulting low-noise field $\Pi_{\tau_K}$ is used as a closure realization. We write this sampling procedure as $\Pi_\theta=G_\theta(V;\zeta)$, where $\zeta$ collects the standard Gaussian variables used for initialization and the subsequent reverse-diffusion steps. 

To represent the persistence of unresolved fluctuations across physical steps, we introduce temporal correlation in these Gaussian variables using stationary Ornstein--Uhlenbeck (OU) processes, following correlated-noise constructions in stochastic parameterization \citep{gagne2020machine}. At successive physical steps of size $\Delta t$, the Gaussian variables evolve as
\begin{equation}\label{eq:ou_sampling_noise}
    \zeta_{\ell+1}=\rho\zeta_\ell+\sqrt{1-\rho^2}\,\widetilde{\zeta}_{\ell+1},
    \qquad
    \rho=\exp\!\left(-\frac{\Delta t}{T_{\mathrm{OU}}}\right),
\end{equation}
where $T_{\mathrm{OU}}$ is the correlation time, and $\zeta_0$ and the newly generated variables $\widetilde{\zeta}_{\ell+1}$ are independent standard Gaussian variables. Separate OU processes are used for initialization and each reverse-diffusion step, with independent realizations for different ensemble members. This construction preserves the standard Gaussian marginal distribution of each noise field while introducing temporal dependence. The nonlinear sampler converts this correlated noise into closure fields using the evolving resolved state.

For fixed Gaussian variables, the sampler is differentiable with respect to its parameters and resolved-state input. We therefore train it through the physical solver using resolved trajectories when reference closure fields are unavailable. Network parameterization, noise settings, and numerical sampling details are given in~\ref{app:implementation}.

\subsection{Indirect training from resolved trajectories}
\label{ssec:closure_training}

We train the closure model from $\mathcal{D}_{\mathrm{state}}$ by coupling the sampler $G_\theta$ introduced in Section~\ref{ssec:conditional_score} to the resolved solver and comparing the simulated trajectories with the recorded states. For each training window, $n\geq2$ trajectories start from the same reference field, $V_\theta^{(m)}(t_0)=V^{\mathrm{ref}}(t_0)$, with $m=1,\ldots,n$. The solver advances from $t_0$ to $t_J$ using $L$ physical steps of size $\Delta t$. At step $\ell$, the current time is $t=t_0+\ell\Delta t$. The recorded times $t_j$ lie on this grid. A closure is sampled at every physical step, so several closure updates occur between consecutive recorded states. Each trajectory uses its current predicted state to sample a closure and advance the resolved dynamics:
\begin{align}
    \Pi_\ell^{(m)}&=G_\theta\!\left(V_\theta^{(m)}(t);\zeta_\ell^{(m)}\right),
    \label{eq:successive_closure_draws}\\
    V_\theta^{(m)}(t+\Delta t)&=\Phi_{t,t+\Delta t}\!\left(V_\theta^{(m)}(t),\Pi_\ell^{(m)}\right).
    \label{eq:indirect_resolved_propagation}
\end{align}
Here, $\Phi_{t,t+\Delta t}$ integrates Eq.~\eqref{eq:hybrid_resolved} over $[t,t+\Delta t)$ with the sampled closure held fixed. The Gaussian variables $\zeta_\ell^{(m)}$ follow Eq.~\eqref{eq:ou_sampling_noise}, with independent OU sequences for different trajectories. Each new closure sample therefore depends on the evolving predicted state and temporally correlated noise, while the reference snapshots provide the targets for training.

We compare the simulated and reference trajectories at the recorded times by stacking their future fields into $Z_\theta^{(m)}=(V_\theta^{(m)}(t_1),\ldots,V_\theta^{(m)}(t_J))$ and $Z^{\mathrm{ref}}=(V^{\mathrm{ref}}(t_1),\ldots,V^{\mathrm{ref}}(t_J))$, respectively. Since the initial field is shared, only future states enter the loss. For each reference window, we estimate the trajectory energy score using the $n$ simulated trajectories \citep{gneiting2007strictly,gneiting2008assessing}:
\begin{equation}\label{eq:trajectory_energy_estimator}
    \mathcal{L}_{\mathrm{ES}}= \frac{1}{n}\sum_{m=1}^{n}\left\|Z_\theta^{(m)}-Z^{\mathrm{ref}}\right\|_{\mathrm{tr}}- \frac{1}{2n(n-1)}\sum_{m\neq m'}\left\|Z_\theta^{(m)}-Z_\theta^{(m')}\right\|_{\mathrm{tr}}.
\end{equation}
Here, $\|\cdot\|_{\mathrm{tr}}$ denotes the Euclidean norm over the stacked fields. The first term measures agreement with the reference trajectory, while the second accounts for variability among simulated trajectories and discourages ensemble collapse. Each distance includes all future snapshots together, allowing the loss to assess their joint distribution. Training averages this loss term over the reference windows. Because different ensemble members use independent OU sequences, Eq.~\eqref{eq:trajectory_energy_estimator} is an unbiased estimator of the trajectory energy score. With the fixed coefficient $1/2$, minimizing the expected energy score over reference trajectories is equivalent to minimizing the energy distance between the predicted and reference trajectory distributions. This distance is defined through expected pairwise distances and equals zero only when the two distributions agree \citep{gneiting2007strictly,szekely2013energy}. The energy-distance relation and the unbiasedness of the ensemble estimator are derived in~\ref{app:training_details}.

The trajectory energy score supervises the resolved evolution without directly observing the closure fields. To control the spatial roughness of these indirectly learned fields, we supplement it with a squared-Laplacian penalty, following standard derivative-based regularization for inverse problems \citep{hansen2010discrete}:
\begin{equation}\label{eq:closure_spatial_regularization}
    \mathcal{R}_{\mathrm{sp}}= \frac{1}{nL}\sum_{m=1}^{n}\sum_{\ell=0}^{L-1}\left\|\ell_f^2\nabla^2\left(\frac{\Pi_\ell^{(m)}}{\Pi_{\mathrm{scale}}}\right)\right\|_{\mathrm{RMS}}^2,
\end{equation}
where $\|\cdot\|_{\mathrm{RMS}}^2$ denotes the spatial average of the squared field. The fixed length scale $\ell_f$ is associated with the spatial filter. The normalization constant $\Pi_{\mathrm{scale}}$ is obtained by dividing the standard deviation of the resolved training states by a fixed reference time, giving it the units of the closure tendency. The penalty is evaluated separately for each sampled closure and suppresses high-wavenumber oscillations more strongly than large-scale variations, without requiring reference closure fields. The total training loss is
\begin{equation}\label{eq:indirect_training_objective}
    \mathcal{L}_{\mathrm{train}} = \mathcal{L}_{\mathrm{ES}}+ \lambda_{\mathrm{sp}}\mathcal{R}_{\mathrm{sp}},
\end{equation}
where $\lambda_{\mathrm{sp}}\geq0$ controls the regularization strength. 

We estimate gradients of the total loss by differentiating through the reverse-diffusion sampler and the integration of the resolved dynamics, including the dependence of later closure samples on previously predicted states. The Gaussian variables defining each OU sequence are held fixed during each gradient evaluation, allowing pathwise differentiation through the sampling procedure \citep{kingma2014auto,rezende2014stochastic}. This trains the closure model through its effects on the resolved evolution, without requiring reference closure fields or time derivatives of the recorded data. Only the closure-model parameters $\theta$ are updated; the physical solver and OU correlation time remain fixed. To reduce the cost of differentiating through repeated closure sampling, we use a randomized discrete adjoint with inverse-probability weighting \citep{oktay2021randomized}. The adjoint is propagated through every step of the resolved solver, while sampler derivatives with respect to both the parameters and the conditioning state are evaluated at independently selected physical steps and weighted by the inverse selection probability. This yields an unbiased estimate of the total-loss gradient at reduced computational cost, while retaining the complete forward simulation.

Algorithm~\ref{alg:indirect_training} summarizes the training procedure. Solver details are documented in~\ref{app:solver}, and numerical training settings are given in~\ref{app:implementation}.

\begin{algorithm}[!htp]
\caption{Indirect training of the stochastic closure}
\label{alg:indirect_training}
\small
\begin{algorithmic}[1]
\REQUIRE $\mathcal{D}_{\mathrm{state}},\Phi,G_\theta,
         \Delta t,L,n\ge2,
         T_{\mathrm{OU}},\lambda_{\mathrm{sp}},\eta$
\STATE Initialize $\theta$
\REPEAT
    \STATE Sample $(V^{\mathrm{ref}}(t_0),\ldots,
                    V^{\mathrm{ref}}(t_J))
                    \sim\mathcal{D}_{\mathrm{state}}$
    \FOR{$m=1,\ldots,n$}
        \STATE $V_\theta^{(m)}(t_0)
                \gets V^{\mathrm{ref}}(t_0)$
        \STATE Generate $\{\zeta_\ell^{(m)}\}_{\ell=0}^{L-1}$
               using Eq.~\eqref{eq:ou_sampling_noise},
               independently of other trajectories
        \FOR{$\ell=0,\ldots,L-1$}
            \STATE $t\gets t_0+\ell\Delta t$
            \STATE $\Pi_\ell^{(m)}
                    \gets G_\theta
                    (V_\theta^{(m)}(t);
                     \zeta_\ell^{(m)})$
            \STATE $V_\theta^{(m)}(t+\Delta t)
                    \gets\Phi_{t,t+\Delta t}
                    (V_\theta^{(m)}(t),
                     \Pi_\ell^{(m)})$
        \ENDFOR
        \STATE $Z_\theta^{(m)}
                \gets(V_\theta^{(m)}(t_1),\ldots,
                       V_\theta^{(m)}(t_J))$
    \ENDFOR
    \STATE Compute $\mathcal{L}_{\mathrm{ES}}$
           using Eq.~\eqref{eq:trajectory_energy_estimator}
    \STATE Compute $\mathcal{R}_{\mathrm{sp}}$
           using Eq.~\eqref{eq:closure_spatial_regularization}
    \STATE $\mathcal{L}_{\mathrm{train}}
            \gets\mathcal{L}_{\mathrm{ES}}
            +\lambda_{\mathrm{sp}}\mathcal{R}_{\mathrm{sp}}$
    \STATE Estimate $\nabla_\theta\mathcal{L}_{\mathrm{train}}$
           using the randomized discrete adjoint
    \STATE Update $\theta$ using AdamW with learning rate $\eta$
\UNTIL{convergence}
\end{algorithmic}
\end{algorithm}

\subsection{Score-based data assimilation of stochastic closures}
\label{ssec:method_closure_da}

After offline training, the learned closure distribution serves as a prior for incorporating observations during prediction. Measurements of the resolved system provide information about closures through their effects on the subsequent evolution, enabling assimilation at the model-form uncertainty level. At physical time $t_j$, we consider observations of the form
\begin{equation}\label{eq:method_observation}
    y_j = \mathcal{H}_j[V^{\mathrm{ref}}(t_j)] + \epsilon_j, \qquad \epsilon_j\sim\mathcal{N}(0,R_j),
\end{equation}
where $\mathcal{H}_j$ maps the resolved field to the measured quantity and $R_j$ is the observation-error covariance. The observation operator can represent partial state measurements or nonlinear functions of the resolved field.

Between consecutive observations, the resolved solver advances through several physical time steps, each supplied with a sampled closure. The observation at $t_j$ therefore constrains the closures through their accumulated effects on the resolved evolution. For an incoming resolved state $V$ at $t_{j-1}$, we collect the closures over these $L$ steps as $\boldsymbol{\Pi}=(\Pi_0,\ldots,\Pi_{L-1})$. Their predicted observation is
\begin{equation}\label{eq:method_closure_observation_map}
    h_j(\boldsymbol{\Pi};V) = \mathcal{H}_j\left[ \Phi_{t_{j-1},t_j}(V,\boldsymbol{\Pi})\right],
\end{equation}
where $\Phi_{t_{j-1},t_j}$ advances the resolved dynamics using the corresponding closure at each physical step. The prior $p_\theta(\boldsymbol{\Pi}\mid V)$ is induced by the learned conditional closure model and its Ornstein--Uhlenbeck (OU)-correlated sampling inputs. The likelihood and target closure posterior are
\begin{align}
    p(y_j\mid\boldsymbol{\Pi},V) &\propto\exp\left[-\frac{1}{2}\left\|y_j-h_j(\boldsymbol{\Pi};V)\right\|_{R_j^{-1}}^2\right],
    \label{eq:method_closure_likelihood}\\
    p_\theta(\boldsymbol{\Pi}\mid V,y_j)&\propto p_\theta(\boldsymbol{\Pi}\mid V)\,p(y_j\mid\boldsymbol{\Pi},V),
    \label{eq:method_closure_posterior}
\end{align}
where $\|\cdot\|_{R_j^{-1}}$ denotes the Euclidean norm weighted by the inverse observation-error covariance $R_j^{-1}$. The prior describes the closure realizations generated by the model, while the likelihood assesses their consistency with the measured evolution.

Score-based sampling incorporates this likelihood information by supplementing the learned prior score with likelihood guidance \citep{song2021score,chung2023diffusion,rozet2023score}. At a noisy diffusion time, evaluating the exact likelihood requires averaging over the clean closures compatible with the noisy intermediate fields. To avoid this high-dimensional conditional expectation, we approximate this likelihood using denoised closure estimates. A prior forecast from the incoming state supplies conditioning fields $\{\widetilde{V}_\ell\}_{\ell=0}^{L-1}$, which remain fixed during the closure update. At each physical step, we obtain a denoised closure estimate, denoted by $D_\theta$, using Tweedie's formula \citep{efron2011tweedie}:
\begin{equation}\label{eq:method_denoised_closure}
    D_\theta(\Pi_{\ell,\tau},\tau,\widetilde{V}_\ell) = \Pi_{\ell,\tau} +\sigma^2(\tau)s_\theta(\Pi_{\ell,\tau},\tau,\widetilde{V}_\ell).
\end{equation}
Writing $\widehat{\boldsymbol{\Pi}}_\tau$ for the resulting denoised closure fields, the approximate observation likelihood is
\begin{equation}\label{eq:method_guidance_likelihood}
    \widetilde{p}(y_j\mid\boldsymbol{\Pi}_\tau,V)=\mathcal{N}\left(y_j;h_j(\widehat{\boldsymbol{\Pi}}_\tau;V),R_j\right).
\end{equation}

At diffusion time $\tau_k$, likelihood guidance uses the gradient of the approximate log-likelihood with respect to the noisy closure at physical step $\ell$:
\begin{equation}\label{eq:method_backward_likelihood_guidance}
    g^{\mathrm{lik}}_{\ell,k}=\nabla_{\Pi_{\ell,\tau_k}}\log\widetilde{p}(y_j\mid\boldsymbol{\Pi}_{\tau_k},V).
\end{equation}
For each evaluation, we advance the resolved solver from the same incoming state using the denoised closure estimates to obtain the predicted observation at $t_j$. We compute these gradients by applying the chain rule through the solver steps and Tweedie's formula, including the derivative of the score-network output with respect to its noisy input. This allows the observation at $t_j$ to guide closure sampling at every physical step within the interval.

To account for temporal correlation in the sampling noise, we weight these likelihood gradients by its covariance matrix $\mathcal{K}^{(j)}$ within observation interval $j$. The sampling update in Eq.~\eqref{eq:reverse_variance_step} then becomes
\begin{equation}\label{eq:method_posterior_score}
    \begin{aligned} 
        \Pi_{\ell,\tau_{k+1}}=\Pi_{\ell,\tau_k}+\Delta\sigma_k^2\left[s_\theta(\Pi_{\ell,\tau_k},\tau_k,\widetilde{V}_\ell)+\sum_{q=0}^{L-1}\mathcal{K}^{(j)}_{\ell q}\,g^{\mathrm{lik}}_{q,k}\right]+\sqrt{\Delta\sigma_k^2}\,\xi_{\ell,k},
    \end{aligned}
\end{equation}
where $\xi_{\ell,k}$ are temporally correlated Gaussian noise fields. Because these perturbations are coupled across physical steps with covariance $\mathcal{K}^{(j)}$, conditioning the reverse diffusion on an observation shifts the drift by the noise covariance times the conditional log-likelihood gradient, consistent with Doob's $h$-transform \citep{chopin2023computational}. We approximate this likelihood gradient via the denoised closures while leaving the prior-score term $s_\theta$ unchanged.

After sampling, we advance the resolved solver from the same incoming state using the updated closures to obtain the forecast at $t_j$. We denote the resulting approximate posterior closure sampler by $G_{\theta,j}^{\mathrm{post}}(V;y_j,R_j,\zeta)$, where $\zeta$ collects the Gaussian sampling inputs. Numerical sampling and guidance settings are given in~\ref{app:da_implementation}.

\subsection{Ensemble Kalman assimilation of resolved states}
\label{ssec:method_ensemble_da}

Ensemble simulations used for uncertainty quantification also provide the sample statistics needed for ensemble Kalman assimilation. The EnKF updates a chosen variable using its cross-covariance with predicted observations, allowing the updated variable to differ from the measured quantity \citep{evensen2003ensemble,burgers1998analysis}. Naturally, we can combine state EnKF with posterior-score DA through progressive correction, using observations to refine closures at the model-form uncertainty level and further correct resolved states at the prediction level. Closure EnKF is also included as a baseline for comparison with posterior-score closure assimilation.

Let $\{X^{f,(m)}\}_{m=1}^{M}$ denote the ensemble to be updated and $\{\widehat{y}_j^{(m)}\}_{m=1}^{M}$ its predicted observations. The superscripts $f$ and $a$ denote forecast and analysis quantities, respectively. The EnKF analysis is
\begin{equation}\label{eq:method_enkf_analysis}
    X^{a,(m)} = X^{f,(m)} + K_X \left[ y_j+\epsilon_j^{(m)} - \widehat{y}_j^{(m)} \right], \qquad K_X = C_{Xh}(C_{hh}+R_j)^{-1},
\end{equation}
where $\epsilon_j^{(m)}\sim\mathcal{N}(0,R_j)$ are independent observation perturbations. The sample covariances are
\begin{align}
    C_{Xh} &= \frac{1}{M-1} \sum_{m=1}^{M} \left(X^{f,(m)}-\overline{X}^{f}\right) \left(\widehat{y}_j^{(m)}-\overline{y}_j\right)^{\mathsf T}, \label{eq:method_cross_covariance}\\
    C_{hh} &= \frac{1}{M-1} \sum_{m=1}^{M} \left(\widehat{y}_j^{(m)}-\overline{y}_j\right) \left(\widehat{y}_j^{(m)}-\overline{y}_j\right)^{\mathsf T}, \label{eq:method_observation_covariance}
\end{align}
with $\overline{X}^{f}$ and $\overline{y}_j$ denoting the corresponding ensemble means. 

For state EnKF, the updated variable is the forecast resolved field, $X^{f,(m)}=V_j^{f,(m)}$, and $\widehat{y}_j^{(m)}=\mathcal{H}_j[V_j^{f,(m)}]$. The cross-covariance $C_{Vh}$ allows observations to correct the full resolved field, including locations where measurements are unavailable. This analysis acts directly at the prediction level, producing the updated state $V_j^{a,(m)}$ from which the next interval begins. For the closure EnKF baseline, we jointly update the closure fields over the observation interval, collected as $X^{f,(m)}=\boldsymbol{\Pi}_j^{f,(m)}$, using the predicted observation $\widehat{y}_j^{(m)}=h_j(\boldsymbol{\Pi}_j^{f,(m)};V_{j-1}^{a,(m)})$. The cross-covariance $C_{\boldsymbol{\Pi}h}$ transfers information from the measured resolved evolution to the closure ensemble. After the analysis, each member is propagated again from its incoming state using the updated closures.

In the combined method, posterior-score DA first conditions the closures, which are then propagated to obtain the forecast for state EnKF. To distribute observation information between these stages, we use progressive correction \citep{musso2001improving,frei2013}, allocating a fraction $0<\alpha_{\mathrm{DA}}<1$ of the observation log-likelihood to posterior-score DA and the remaining fraction to state EnKF. For Gaussian observation errors, this allocation is implemented through
\begin{equation}\label{eq:method_likelihood_allocation}
    R_{\Pi,j}=\frac{R_j}{\alpha_{\mathrm{DA}}}, \qquad R_{V,j}=\frac{R_j}{1-\alpha_{\mathrm{DA}}}.
\end{equation}
Standalone closure or state assimilation uses the original covariance $R_j$. The combined procedure therefore shares the observation information between closure refinement and state correction without applying the full likelihood at both stages.

Sequential assimilation repeats these updates as observations become available. Each member draws its next closure using its own updated resolved state, while the trained closure model remains fixed. Algorithm~\ref{alg:sequential_da} summarizes posterior-score DA for the closure term, together with EnKF for the system state. Standalone posterior-score DA uses $R_{\Pi,j}=R_j$, while the combined method uses the covariance allocation in Eq.~\eqref{eq:method_likelihood_allocation}. Here, $\operatorname{EnKF}$ denotes the ensemble analysis in Eq.~\eqref{eq:method_enkf_analysis}. Numerical details of ensemble analysis and solver integration are given in~\ref{app:da_implementation} and~\ref{app:solver}.

\begin{algorithm}[!htp]
\caption{Hybrid simulation with posterior-score DA and state EnKF}
\label{alg:sequential_da}
\small
\begin{algorithmic}[1]
\REQUIRE $G_{\theta,j}^{\mathrm{post}},\Phi,
         \{V_0^{a,(m)}\}_{m=1}^{M},
         \{t_j\}_{j=0}^{J},
         \{y_j,\mathcal{H}_j,R_{\Pi,j},R_{V,j}\}_{j=1}^{J}$
\FOR{$j=1,\ldots,J$}
    \FOR{$m=1,\ldots,M$}
        \STATE Generate Gaussian sampling inputs
               $\zeta_j^{(m)}$
        \STATE $\boldsymbol{\Pi}_j^{(m)}
                \gets G_{\theta,j}^{\mathrm{post}}
                (V_{j-1}^{a,(m)};
                 y_j,R_{\Pi,j},\zeta_j^{(m)})$
        \STATE $V_j^{f,(m)}
                \gets\Phi_{t_{j-1},t_j}
                (V_{j-1}^{a,(m)},\boldsymbol{\Pi}_j^{(m)})$
        \STATE $V_j^{a,(m)}\gets V_j^{f,(m)}$
    \ENDFOR
    \IF{state EnKF is used}
        \STATE $\{V_j^{a,(m)}\}_{m=1}^{M}
                \gets\operatorname{EnKF}
                (\{V_j^{f,(m)}\}_{m=1}^{M},
                 \{\mathcal{H}_j[V_j^{f,(m)}]\}_{m=1}^{M};
                 y_j,R_{V,j})$
    \ENDIF
\ENDFOR
\end{algorithmic}
\end{algorithm}

\section{Results}\label{sec:results}

\subsection{Numerical setup}\label{ssec:numerical_setup}
We evaluate the proposed framework using the two-dimensional incompressible Navier--Stokes equations in vorticity form on the doubly periodic domain $\Omega=(0,L)^2$, with $L=2\pi$:
    \begin{equation}\label{eq:stochastic_ns}
    \begin{aligned}
        d\omega = \bigl[ -\mathbf{u}\cdot\nabla\omega & +\nu\nabla^2\omega -\alpha\omega +f \bigr]dt +\sigma_{\mathrm f}Q^{1/2}dW_t, \\
        \nabla\cdot\mathbf{u}&=0, \qquad \omega(\mathbf{x},0)=\omega_0(\mathbf{x}).
    \end{aligned}
    \end{equation}
Here, $\omega$ denotes the vorticity, $\nu$ the kinematic viscosity, and $\alpha$ the linear damping coefficient. The velocity is recovered from the streamfunction through $\mathbf{u}=(\partial_{y}\psi, -\partial_{x}\psi)$, with $-\Delta\psi=\omega$. The deterministic forcing is prescribed as
    \begin{equation}\label{eq:deterministic_forcing}
        f(\mathbf{x}) = \cos\left(\frac{2\pi k_f}{L}(x+y)\right) + \sin\left(\frac{2\pi k_f}{L}(x+y)\right),
    \end{equation}
where $k_f$ controls the forcing length scale. The training regime
uses $\nu=5\times10^{-4}$, $\alpha=0.1$, and $k_f=6$.

The final term in Eq.~\eqref{eq:stochastic_ns} introduces additive Gaussian forcing that is white in time and spatially correlated. Here, $W_t$ is a Wiener process, $Q$ specifies the spatial covariance, and $\sigma_{\mathrm f}$ controls the forcing amplitude. We take $Q$ to be diagonal in Fourier space, with equal power assigned to modes in 
    \begin{equation}\label{eq:noise_band}
        \mathcal{K}_{\mathrm f} =  \left\{ \mathbf{k}\in(2\pi/L)\mathbb{Z}^2: 48\leq|\mathbf{k}|\leq96 \right\},
    \end{equation}
and zero power outside this band. The band-limited Gaussian fields are normalized to unit expected spatial mean square before forming the stochastic increments. We prescribe a stochastic enstrophy injection rate equal to $4\%$ of the mean deterministic injection measured in a noise-free reference simulation, giving $\sigma_{\mathrm f}\approx0.291$. This introduces weak random forcing at unresolved scales while keeping its prescribed injection well below the deterministic reference level.

The stochastic forcing represents fluctuations at scales below the resolution of the coarse model introduced below. Its support is chosen outside the retained Fourier modes, so its direct contribution vanishes under coarse-graining. These unresolved fluctuations nevertheless influence the resolved flow through nonlinear interactions. Independent forcing realizations therefore introduce variability into the filtered dynamics even when the complete initial state and deterministic forcing are fixed. 

\subsubsection{Filtered dynamics and closure formulation}\label{sssec:filters}

Direct numerical simulations (DNS) of Eq.~\eqref{eq:stochastic_ns} use Fourier pseudospectral discretization in both spatial directions on a grid with $N_{\mathrm{DNS}}=512$ points in each direction \citep{canuto2006spectral}. The deterministic terms are advanced using a second-order semi-implicit scheme with time step $\Delta t=10^{-3}\,\mathrm{s}$, and independent Gaussian forcing increments are added at each step. Numerical details are given in~\ref{app:solver}. For large-eddy simulations (LES), the resolved dynamics are represented on a coarse grid with $N_{\mathrm{LES}}=64$ points in each direction. We obtain the resolved vorticity by applying a Gaussian spatial filter followed by spectral truncation:
\begin{equation}\label{eq:resolved_vorticity}
    \overline{\omega}=\mathcal{P}_{\Delta}\omega=\mathcal{R}_{N_{\mathrm{LES}}}\left(G_{\Delta}*\omega\right),
\end{equation}
where $\mathcal{R}_{N_{\mathrm{LES}}}$ retains the Fourier modes represented on the LES grid, and $*$ denotes periodic convolution. The Gaussian filter has the Fourier transfer function
\begin{equation}\label{eq:gaussian_filter}
    \widehat{G}_{\Delta}(\mathbf{k})=\exp\left(-\frac{\Delta^2|\mathbf{k}|^2}{24}\right), \qquad \Delta=4\frac{L}{N_{\mathrm{LES}}}.
\end{equation}
We similarly define $\overline{\mathbf{u}}=\mathcal{P}_{\Delta}\mathbf{u}$ and $\overline{f}=\mathcal{P}_{\Delta}f$. The resulting filtered DNS (FDNS) fields serve as the reference solutions for evaluating the coarse models.

\begin{figure}[!htb]
    \centering
    \includegraphics[width=\linewidth]{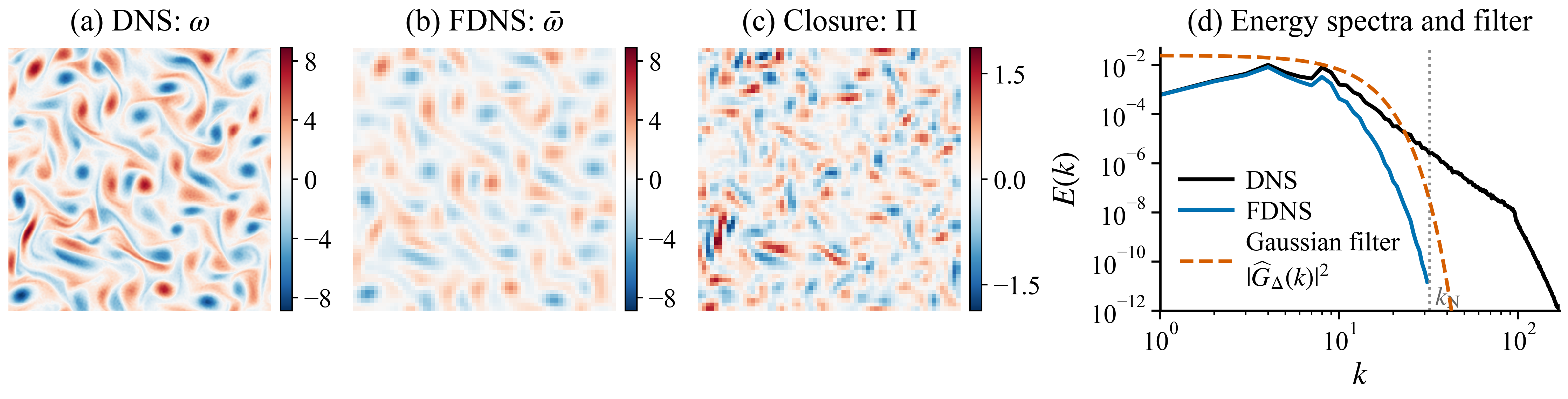}
    \caption{Spatial filtering and closure formation: (a) DNS vorticity $\omega$ on the $512\times512$ grid; (b) filtered vorticity $\overline{\omega}$ on the $64\times64$ resolved grid; (c) the corresponding subfilter-scale closure $\Pi$, defined in Eq.~\eqref{eq:closure_definition}; and (d) kinetic energy spectra $E(k)$ of the DNS and FDNS fields. The dashed curve shows the squared filter transfer function. The vertical dotted line marks the LES cutoff at $k_{\mathrm{N}}=32$.}
    \label{fig:filtering_closure}
\end{figure}

By construction of the forcing band in Eq.~\eqref{eq:noise_band}, $\mathcal{P}_{\Delta}Q^{1/2}=0$ since the stochastic forcing is supported only outside the retained Fourier modes. Applying the coarse-graining operator $\mathcal{P}_{\Delta}$ to Eq.~\eqref{eq:stochastic_ns} therefore eliminates the explicit stochastic forcing term and gives
\begin{equation}\label{eq:filtered_unclosed}
    \frac{\partial\overline{\omega}}{\partial t}=-\mathcal{P}_{\Delta}\left(\mathbf{u}\cdot\nabla\omega\right)+\nu\nabla^2\overline{\omega}-\alpha\overline{\omega}+\overline{f}.
\end{equation}
In Eq.~\eqref{eq:filtered_unclosed}, the nonlinear advection term depends on fine-scale fields that are unavailable to the coarse solver, so the equation is not closed in the resolved variables. We isolate this unresolved contribution by defining the subfilter-scale closure as the difference between the filtered fine-scale advection and the advection evaluated from the resolved fields:
\begin{equation}\label{eq:closure_definition}
    \Pi=\mathcal{P}_{\Delta}\left(\mathbf{u}\cdot\nabla\omega\right)-\overline{\mathbf{u}}\cdot\nabla\overline{\omega}.
\end{equation}
Substituting Eq.~\eqref{eq:closure_definition} into Eq.~\eqref{eq:filtered_unclosed} yields
\begin{equation}\label{eq:resolved_dynamics}
    \frac{\partial\overline{\omega}}{\partial t}=-\overline{\mathbf{u}}\cdot\nabla\overline{\omega}+\nu\nabla^2\overline{\omega}-\alpha\overline{\omega}+\overline{f}-\Pi.
\end{equation}
The closure $\Pi$ accounts for the effects of unresolved nonlinear interactions on the evolution of the resolved vorticity, including the influence of the stochastically forced unresolved modes. As shown in Eq.~\eqref{eq:closure_definition}, its evaluation requires fine-scale information in $\mathbf{u}$ and $\omega$ that is not uniquely determined by the resolved state. Different fine-scale realizations can therefore yield different closure fields for the same resolved field. Figure~\ref{fig:filtering_closure} illustrates this coarse-graining through a representative DNS snapshot, its filtered counterpart, and the corresponding closure field. The resolved vorticity retains the larger flow structures while smoothing the fine-scale filaments, with the spectral comparison showing the associated attenuation of high-wavenumber energy.

A deterministic closure assigns a single correction to each resolved state and cannot represent this conditional variability. We therefore model the closure as a conditional random field $\Pi_{\theta}\sim p_{\theta}\left(\Pi\mid\overline{\omega}\right)$. In coarse simulations, samples from this distribution replace $\Pi$ in Eq.~\eqref{eq:resolved_dynamics}, retaining the resolved dynamics while providing a stochastic representation of the unresolved interactions.

\subsubsection{Training data and comparative evaluation}\label{sssec:data_evaluation}

We generate $80$ independent DNS trajectories under the training configuration described above. Each simulation undergoes a $60\,\mathrm{s}$ deterministic spin-up, followed by $100\,\mathrm{s}$ with stochastic forcing active to reach statistical equilibrium before data collection. We then record the resolved vorticity and corresponding analytic closure fields every $0.04\,\mathrm{s}$ over a $40\,\mathrm{s}$ interval, yielding $1000$ paired snapshots per trajectory. The trajectories are divided into training, validation, and test sets containing $64$, $8$, and $8$ trajectories, respectively. This partition is performed at the trajectory level, so snapshots from the same realization remain within a single subset.

For indirect training, we retain the full resolved fields but subsample the trajectories at intervals of $0.2\,\mathrm{s}$, resulting in a training set with $200$ resolved snapshots per trajectory. Training uses short windows comprising an initial state and two subsequent observations, spanning $0.4\,\mathrm{s}$. Starting from the initial state, the coarse solver advances the resolved dynamics using sampled closures, and the generated trajectories are compared with the reference observations through the trajectory energy score loss. Reference closure fields are not used in this procedure.

We denote the indirectly trained stochastic closure model by Score-I and compare it with its directly trained counterpart, Score-D. Both couple a conditional score-based closure model to the resolved physical solver, forming hybrid models. Whereas Score-I is trained using only resolved-state trajectories, Score-D is trained through denoising score matching using all $64{,}000$ paired resolved states and analytic closures. In contrast to the hybrid models, we also consider two purely data-driven black-box predictive models: a Fourier neural operator (FNO) and a conditional state diffusion model (State-DM). The FNO predicts the resolved-state increment deterministically, whereas State-DM learns its conditional distribution. Both use the same resolved trajectories sampled at $0.2\,\mathrm{s}$ intervals as Score-I and advance the state through a learned transition map. The filtered deterministic forcing, viscosity, and damping coefficient are supplied as additional inputs, giving these predictors access to the prescribed physical parameters for fair comparisons. In addition, a dynamic Smagorinsky model (DSM) is also used in some comparisons to provide a conventional deterministic closure baseline.

The numerical results are organized into three parts, progressing from indirect closure learning to generalization across physical regimes and online data assimilation:

\begin{itemize}
    \item We first perform in-distribution validation of the stochastic closures learned indirectly from resolved trajectories in Section~\ref{ssec:indirect_validation}. Comparisons with reference closure fields, together with coarse-grid simulations, demonstrate that Score-I captures physically relevant closure structures and improves resolved-flow predictions without using closure fields during training.

    \item We next examine out-of-distribution generalization by comparing hybrid and purely data-driven models under changes in forcing and viscosity. These experiments demonstrate the greater robustness of hybrid models when the physical regime departs from the training conditions, as presented in Section~\ref{ssec:generalization}.

    \item We then consider simulations with online observations available over time. Compared with closure EnKF, score-based data assimilation more accurately captures non-Gaussian closure statistics and reduces state prediction errors in sequential assimilation. We further show that combining closure conditioning with state EnKF through progressive correction provides additional improvements in state estimation. These results are presented in Section~\ref{ssec:data_assimilation}.
\end{itemize}

To evaluate these experiments consistently, we introduce common metrics for prediction accuracy and ensemble variability. We report prediction accuracy using the relative $L^2$ error (RE). The root mean square error (RMSE) is also defined for comparison with ensemble spread when assessing uncertainty calibration:
\begin{equation}\label{eq:prediction_errors}
    \mathrm{RE}(z,z^{\mathrm{ref}})=\frac{\|z-z^{\mathrm{ref}}\|_{\mathrm{F}}}{\|z^{\mathrm{ref}}\|_{\mathrm{F}}}, \qquad \mathrm{RMSE}(z,z^{\mathrm{ref}})=\frac{\|z-z^{\mathrm{ref}}\|_{\mathrm{F}}}{\sqrt{N_g}},
\end{equation}
where $z$ and $z^{\mathrm{ref}}$ denote the predicted and reference closure or resolved vorticity fields on a common grid of $N_g$ points, and $\|\cdot\|_{\mathrm{F}}$ is the Frobenius norm.

For an ensemble $\{z^{(m)}\}_{m=1}^{M}$, forecast errors are evaluated using the ensemble mean
\begin{equation}\label{eq:ensemble_mean}
    \mu_z=\frac{1}{M}\sum_{m=1}^{M}z^{(m)}.
\end{equation}
The ensemble spread is the square root of the spatially averaged sample variance:
\begin{equation}\label{eq:ensemble_spread}
    S_z=\left[\frac{1}{N_g(M-1)}\sum_{m=1}^{M}\|z^{(m)}-\mu_z\|_{\mathrm{F}}^2\right]^{1/2}, \qquad M>1.
\end{equation}
The ratio $\mathrm{RMSE}(\mu_z,z^{\mathrm{ref}})/S_z$ provides an aggregate diagnostic of uncertainty calibration, with values near $1$ indicating that the ensemble spread is consistent with the magnitude of the ensemble-mean prediction error.

We also examine the resolved kinetic energy and enstrophy to characterize the physical behavior of the simulated flow:
\begin{equation}\label{eq:resolved_energy_enstrophy}
    \mathcal{E}(t)=\frac{1}{2|\Omega|}\int_{\Omega}|\overline{\mathbf{u}}(\mathbf{x},t)|^2\,\mathrm{d}\mathbf{x}, \qquad \mathcal{Z}(t)=\frac{1}{2|\Omega|}\int_{\Omega}\overline{\omega}(\mathbf{x},t)^2\,\mathrm{d}\mathbf{x}.
\end{equation}
The kinetic energy spectrum $E(k,t)$ is computed from velocity Fourier energies in radial wavenumber bins, satisfying $\sum_k E(k,t)\Delta k=\mathcal{E}(t)$ with $\Delta k=2\pi/L$. For ensembles, all three quantities are computed for each member before averaging.

\subsection{Validation of indirectly learned stochastic closures}\label{ssec:indirect_validation}

We evaluate Score-I through a priori closure diagnostics and a posteriori coarse-grid simulations. As a directly supervised baseline, we implement Score-D using paired resolved states and analytically evaluated closure fields, following the methodology in \cite{dong2025data}. Both models share the same FNO architecture and number of trainable parameters. They are also trained on the same 64 FDNS trajectories: Score-D uses state--closure pairs recorded every $0.04\,\mathrm{s}$, whereas Score-I uses resolved states sampled every $0.2\,\mathrm{s}$ without closure labels. Score-I therefore has access to only one fifth as many resolved snapshots and must infer closure effects through their influence on subsequent flow evolution. 

\subsubsection{A priori analysis of learned closures}\label{sssec:apriori_closures}

\begin{figure}[!htb]
    \centering
    \includegraphics[width=\textwidth]{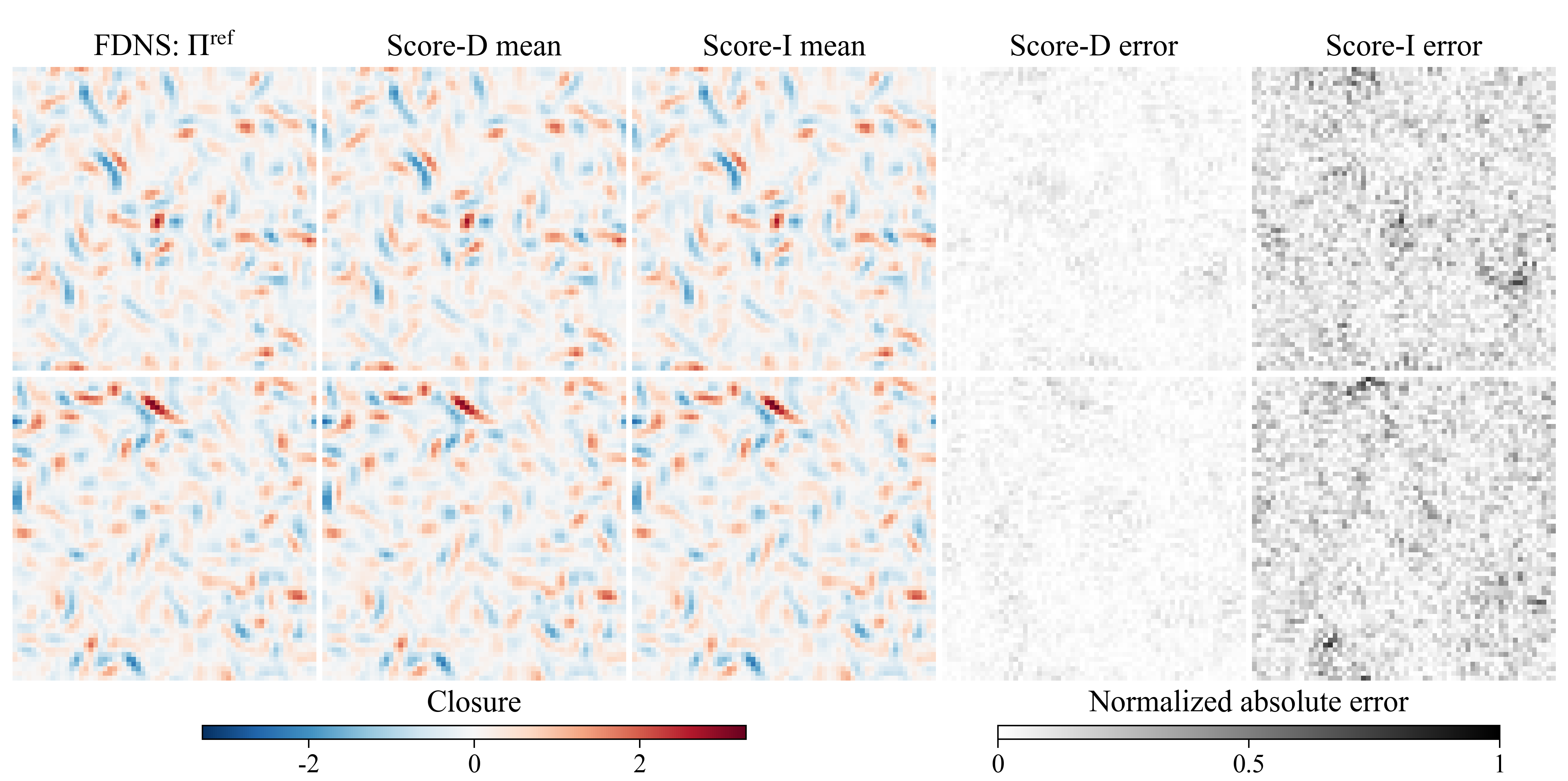}
    \caption{A priori comparison of closure fields at two resolved states. Each row shows, from left to right, the FDNS reference closure $\Pi$, the Score-D and Score-I ensemble means, and their corresponding prediction errors relative to FDNS. Each ensemble mean is obtained from $32$ closure samples conditioned on the same resolved state.}
    \label{fig:closure_samples}
\end{figure}

We begin with an a priori evaluation of Score-I against Score-D, examining sample accuracy, physical effects, and distributional agreement. Single-sample errors are averaged over held-out resolved states, with one closure draw per state. Ensemble-mean errors are evaluated by averaging 32 closure samples conditioned each selected state before computing the relative error and averaging over all 64 selected states. Table~\ref{tab:closure_sampling_error} shows that Score-D achieves lower errors for both individual samples and ensemble means, consistent with its direct supervision using more densely sampled state--closure pairs. Ensemble averaging reduces the errors for both models. Figure~\ref{fig:closure_samples} further compares the reference closure fields with ensemble means derived from 32 samples per model and their normalized absolute errors. Although Score-I exhibits larger local discrepancies, its ensemble means recover the dominant spatial structures of the reference closures, demonstrating that recognizable closure structures can be recovered effectively from resolved trajectories despite the more limited supervision. These errors measure agreement with the available reference closure realization. The spectral and density comparisons below complement this assessment by examining the physical effects and statistical variability of the generated closures.

\begin{table}[!htb]
    \centering
    \setlength{\tabcolsep}{12pt}
    \renewcommand{\arraystretch}{1.15}
    \caption{Relative errors of Score-D and Score-I closure predictions against paired FDNS closures at held-out resolved states. Columns report errors for individual closure samples and ensemble means. Bold values indicate the lowest error in each column.}
    \label{tab:closure_sampling_error}
    \begin{tabular}{@{}lcc@{}}
        \toprule
        Model & Single sample & Ensemble mean \\
        \midrule
        Score-D & \textbf{6.815e-02} & \textbf{4.335e-02} \\
        Score-I & 3.642e-01 & 1.517e-01 \\
        \bottomrule
    \end{tabular}
\end{table}

\begin{figure}[!htb]
    \centering
    \includegraphics[width=\textwidth]{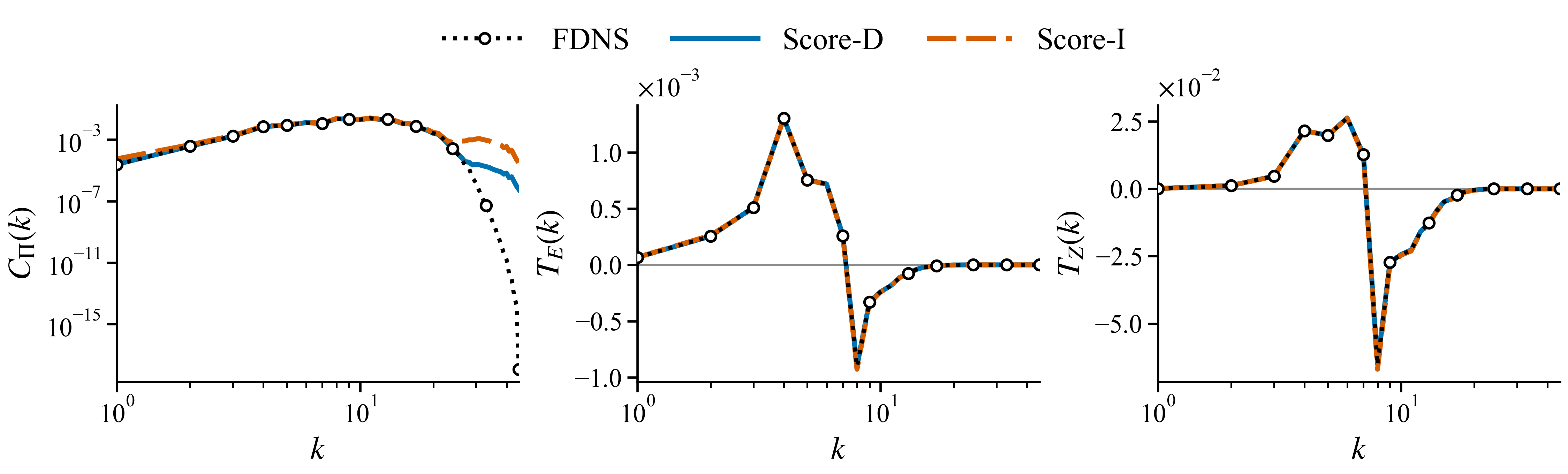}
    \caption{Spectral comparison of FDNS, Score-D, and Score-I closures: closure power spectrum $C_\Pi(k)$, SGS energy transfer $T_E(k)$, and SGS enstrophy transfer $T_Z(k)$, from left to right. Positive and negative transfer values indicate addition to and removal from the resolved energy or enstrophy, respectively.}
    \label{fig:closure_physics}
\end{figure}

The impact of these pointwise closure errors on coarse-grid predictions depends on the scales at which they occur and their influence on the resolved flow. Figure~\ref{fig:closure_physics} thus compares the closure power spectrum $C_\Pi(k)$ and the SGS energy and enstrophy transfer spectra, $T_E(k)$ and $T_Z(k)$. The transfer spectra describe closure contributions to resolved energy and enstrophy at each wavenumber, with positive and negative values indicating addition and removal, respectively. The spectral definitions are provided in~\ref{app:metrics}. Both models reproduce the closure power at low and intermediate wavenumbers, while overpredicting its high-wavenumber tail, particularly for Score-I. Their transfer spectra nevertheless closely follow the reference across both positive and negative transfer regions. Since these transfers depend on correlations between the closure and resolved fields, their agreement shows that Score-I captures the dominant contributions of closures to the resolved energy and enstrophy even when its small-scale closure amplitudes are less accurate.

\begin{figure}[!htb]
    \centering
    \includegraphics[width=\textwidth]{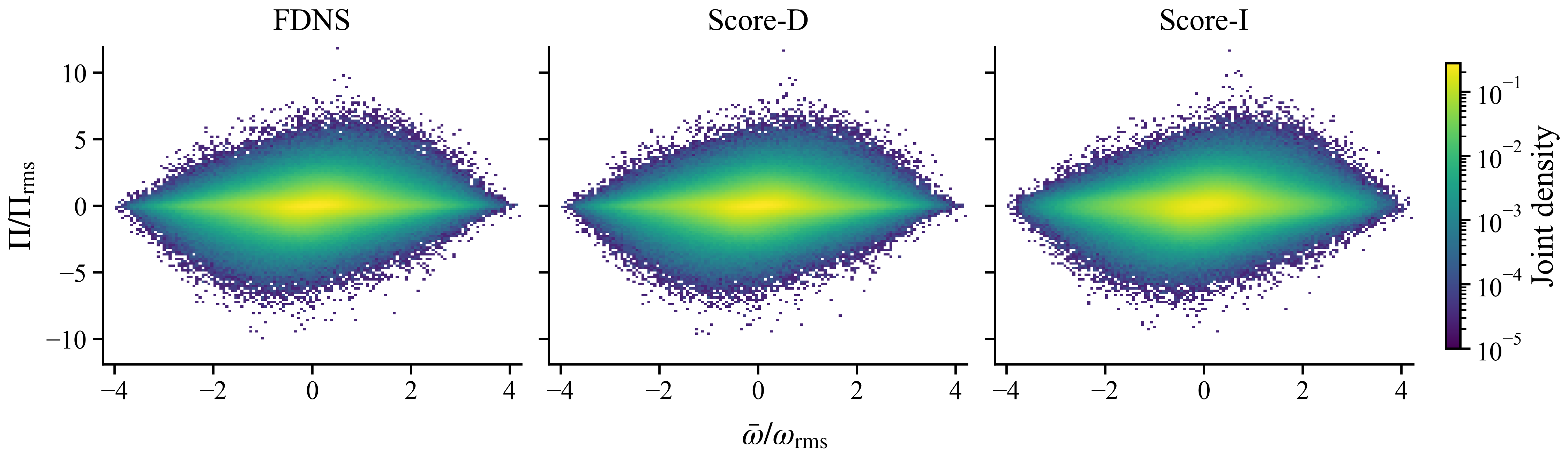}
    \caption{Joint probability densities of resolved vorticity $\overline{\omega}$ and closure $\Pi$ for FDNS, Score-D, and Score-I. Densities are estimated from local state--closure pairs collected across spatial points and held-out resolved states, using individual generated closure samples for the learned models. All panels share a common logarithmic density scale.}
    \label{fig:joint_state_closure}
\end{figure}

We also examine the distribution of closure values and their statistical dependence on local resolved vorticity. Score-D learns this relationship directly from paired fields, while Score-I infers it through resolved trajectories. Figure~\ref{fig:joint_state_closure} compares joint densities of local resolved vorticity and closure values for FDNS, Score-D, and Score-I. Each density is estimated separately using local state--closure pairs collected across spatial points and held-out resolved states. Both models reproduce the main features of the reference density, including its concentration near zero closure and its narrowing at larger vorticity magnitudes. Score-D achieves closer agreement in the detailed density contours, while Score-I produces a slightly broader distribution with less pronounced narrowing.

Together, these comparisons show that Score-I learns the dominant closure structures and their contributions to resolved energy and enstrophy, while Score-D more closely reproduces pointwise values and joint statistics. We next examine how these differences affect predictions in coarse-grid simulations.

\subsubsection{A posteriori validation in coarse-grid simulations}\label{sssec:aposteriori_closures}

We evaluate the learned closures in forecasts without data assimilation from $4$ held-out initial conditions under the training regime, using $512$ ensemble members for each stochastic model. All models use the same coarse-grid solver and initial resolved fields. As a conventional baseline, we include the dynamic Smagorinsky model (DSM), which represents unresolved interactions through a deterministic eddy-viscosity closure. Its coefficient is estimated from the resolved and test-filtered fields through a spatially averaged least-squares fit \citep{germano1991dynamic,lilly1992proposed}.

\begin{figure}[!htb]
    \centering
    \includegraphics[width=\textwidth]{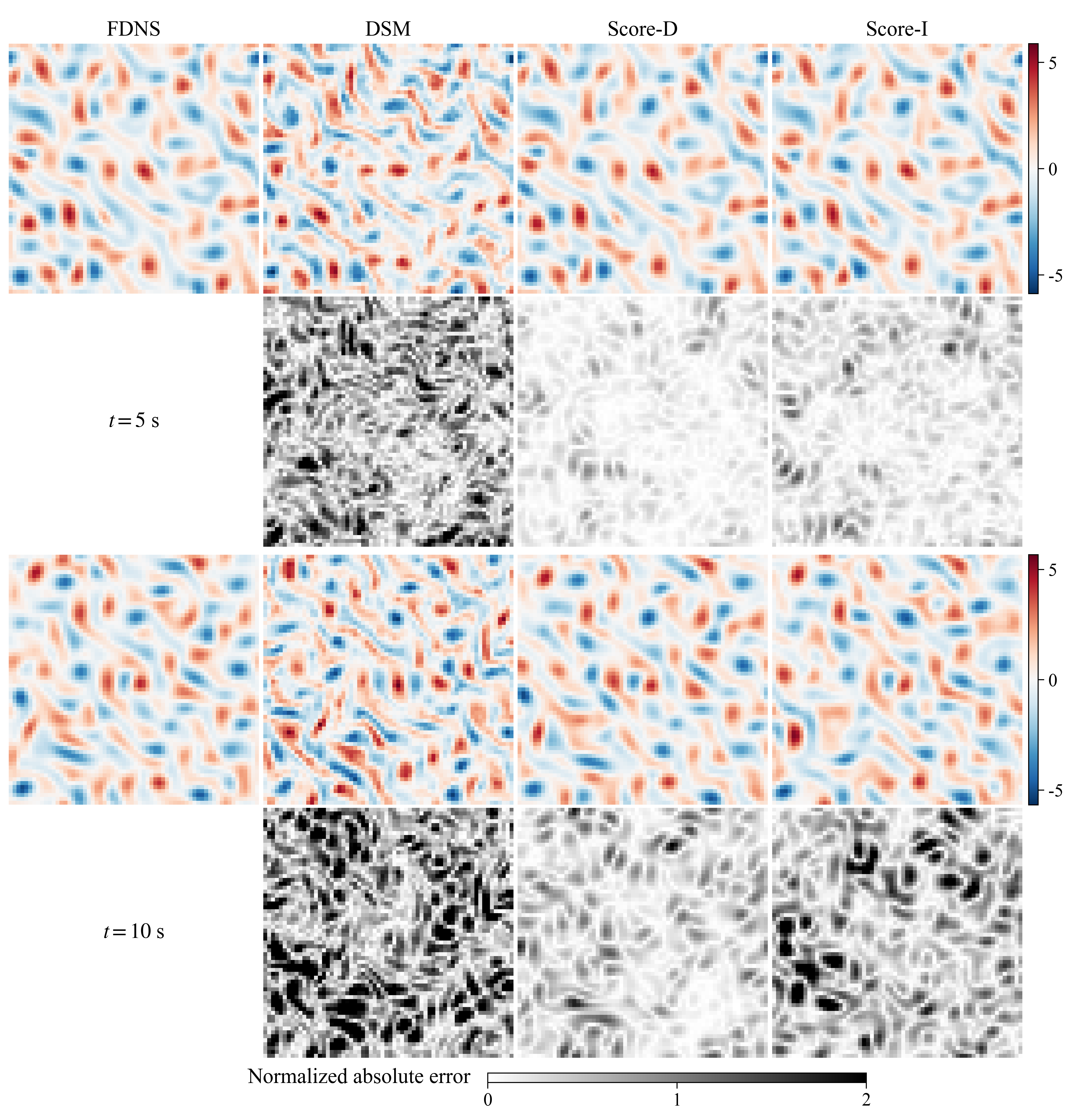}
    \caption{Resolved vorticity predictions using DSM, Score-D, and Score-I at $t=5$ and $10\,\mathrm{s}$ in the training regime. At each time, the vorticity fields are followed by their corresponding prediction errors relative to FDNS. Score-D and Score-I fields show individual ensemble realizations.}
    \label{fig:training_state_fields}
\end{figure}

Figure~\ref{fig:training_state_fields} compares the predicted vorticity fields with the FDNS reference at $t=5$ and $10\,\mathrm{s}$. Both simulations with trained score models preserve the dominant flow structures more closely than DSM, whose predictions develop stronger small-scale distortions. Score-D provides the closest agreement, while Score-I retains the principal structures with larger spatial discrepancies as the forecasts progress.

\begin{figure}[!htb]
    \centering
    \includegraphics[width=\textwidth]{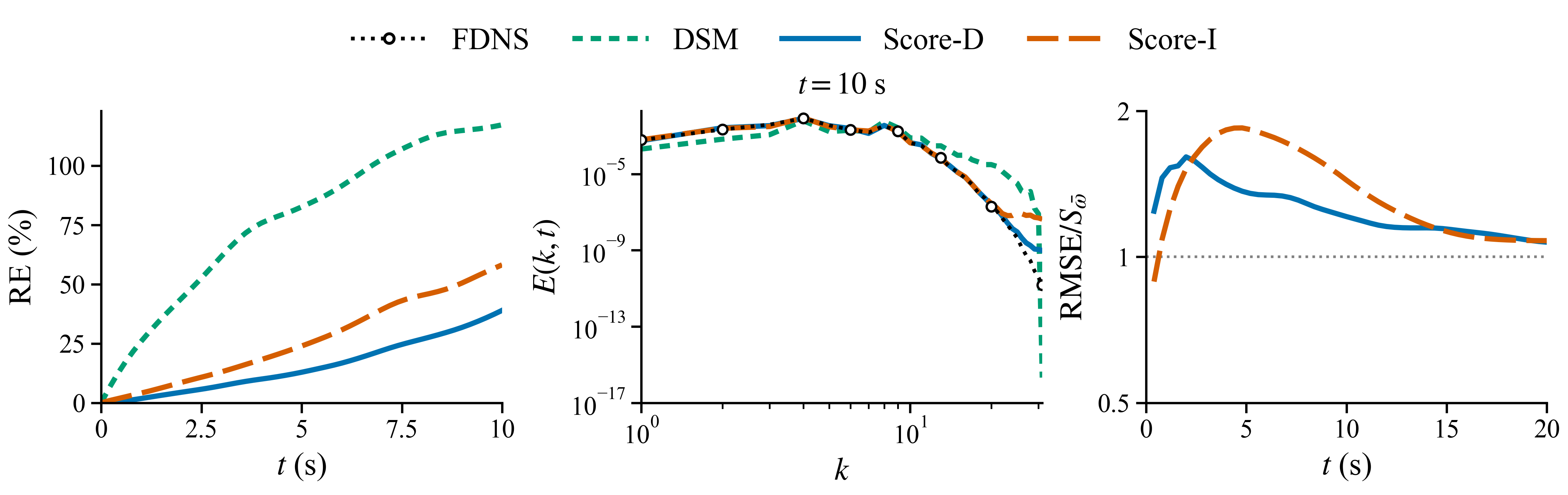}
    \caption{A posteriori diagnostics of DSM, Score-D, and Score-I in the training regime: relative state prediction errors, kinetic energy spectra $E(k)$ at $t=10\,\mathrm{s}$, and RMSE-to-spread ratios, from left to right. Prediction errors use ensemble means for the stochastic models. The dotted horizontal line in the right panel marks equal RMSE and ensemble spread.}
    \label{fig:training_rollout_diagnostics}
\end{figure}

The state prediction errors in Table~\ref{tab:training_state_errors} and the left panel of Figure~\ref{fig:training_rollout_diagnostics} confirm the improvement provided by both learned closures. At $t=10\,\mathrm{s}$, RE decreases from $1.172$ for DSM to $0.390$ for Score-D and $0.582$ for Score-I. The higher errors of Score-I are consistent with the a priori comparison and the more challenging task of inferring closures from sparsely sampled resolved trajectories. Its substantial improvement over DSM nevertheless shows that this indirect supervision produces effective corrections to the coarse-grid dynamics.

\begin{table}[!htb]
    \centering
    \caption{Relative state prediction errors of DSM, Score-D, and Score-I against FDNS in the training regime. Errors for the stochastic models use ensemble-mean predictions. Bold values indicate the lowest error at each time.}
    \label{tab:training_state_errors}
    \setlength{\tabcolsep}{12pt}
    \renewcommand{\arraystretch}{1.15}
    \begin{tabular}{@{}cccc@{}}
        \toprule
        $t$ (s) & DSM & Score-D & Score-I \\
        \midrule
        $2.5$  & 5.259e-01 & \textbf{5.730e-02} & 1.079e-01 \\
        $5.0$  & 8.258e-01 & \textbf{1.288e-01} & 2.394e-01 \\
        $7.5$  & 1.070e+00 & \textbf{2.460e-01} & 4.303e-01 \\
        $10.0$ & 1.172e+00 & \textbf{3.897e-01} & 5.816e-01 \\
        \bottomrule
    \end{tabular}
\end{table}

The improvement in state accuracy is accompanied by closer agreement in the distribution of kinetic energy across spatial scales. The middle panel of Figure~\ref{fig:training_rollout_diagnostics} shows that both score models closely follow the FDNS spectrum over the energy-containing wavenumbers, whereas DSM exhibits a more pronounced excess of energy at higher wavenumbers. Score-I retains more energy in the spectral tail than Score-D, consistent with its larger discrepancies in the high-wavenumber closure power observed in the a priori tests. The learned closures therefore reproduce the principal energy distribution while differing in their accuracy at the smaller resolved scales.

The stochastic closures also generate an ensemble of flow evolutions. The right panel of Fig.~\ref{fig:training_rollout_diagnostics} shows that $\mathrm{RMSE}/S_{\bar{\omega}}$ exceeds unity through much of the early forecast, indicating that the ensemble spread initially underestimates the prediction error. As the ensemble evolves, both ratios approach unity, reaching approximately $1.07$ for Score-D and $1.08$ for Score-I at $t=20\,\mathrm{s}$. This progression indicates improving calibration, as the ensemble spread approaches the magnitude of the ensemble-mean prediction error. Here, all members start from the same resolved state for each test case. The ensemble spread therefore develops as closure variability propagates through the resolved dynamics, representing uncertainty generated progressively during the forecast.

Score-I improves on DSM in state accuracy and spectral agreement, while its ensemble spread approaches the magnitude of forecast error. These benefits persist over forecast intervals substantially longer than the $0.4\,\mathrm{s}$ training windows, showing that closures learned from resolved trajectories remain useful during forecasts without data assimilation. Section~\ref{ssec:generalization} examines how the hybrid models respond to changes in forcing and viscosity.

\subsection{Generalization across physical regimes}\label{ssec:generalization}

The hybrid models combine closure sampling with numerical integration of the resolved equations, retaining an explicit representation of how forcing and dissipation affect the flow. When these physical parameters change, their contributions can be updated directly in the solver while the learned closure remains fixed. We investigate the predictive benefits of retaining this physical structure by comparing hybrid and purely data-driven models under regime changes, starting from comparable short-window prediction accuracy.

The purely data-driven baselines comprise a deterministic Fourier neural operator (FNO) and a conditional state diffusion model (State-DM). Both advance the resolved vorticity over $\Delta t_{\mathrm{pred}}=0.2\,\mathrm{s}$ by learning the state increment $\Delta\overline{\omega}_n=\overline{\omega}_{n+1}-\overline{\omega}_n$. The FNO predicts this increment as
\begin{equation}\label{eq:state_fno_prediction}
    \Delta\overline{\omega}_n=\mathcal{F}_{\phi}\left(\overline{\omega}_n,\overline{f},\nu,\alpha\right),
\end{equation}
whereas State-DM learns its conditional distribution,
\begin{equation}\label{eq:state_diffusion_prediction}
    \Delta\overline{\omega}_n\sim q_{\vartheta}\left(\Delta\overline{\omega}\mid\overline{\omega}_n,\overline{f},\nu,\alpha\right).
\end{equation}
Here, $\phi$ and $\vartheta$ denote the parameters of the FNO and State-DM predictors, respectively. The predicted increment is then added to the current state, and the resulting transition is applied recursively. Both baselines use the same 64 training trajectories sampled every $0.2\,\mathrm{s}$ and receive the prescribed filtered forcing, viscosity, and damping coefficient as inputs. The FNO is trained over two consecutive transitions, while State-DM learns the conditional score of the next-state increment. Neither baseline requires closure labels.

To compare subsequent error accumulation from similar initial accuracy, we hold Score-I fixed and select the other models using training-regime validation errors at $0.2$ and $0.4\,\mathrm{s}$. On the independent test set, the selected models have RE within $20\%$ of the corresponding Score-I error at both forecast times. Note that the Score-D checkpoint used here differs from the best-performing one used in the preceding subsection. Table~\ref{tab:short_window_matching} reports the independent test errors of the selected models. Their comparable short-window accuracy provides a common starting point for the regime-transfer comparisons.

\begin{table}[!htb]
    \centering
    \caption{Short-window relative state prediction errors of Score-D, Score-I, FNO, and State-DM in the training regime. The selected models provide comparable accuracy at $t=0.2$ and $0.4\,\mathrm{s}$ for the subsequent regime-transfer comparisons.}
    \label{tab:short_window_matching}
    \begin{tabular}{lcc}
        \toprule
        Model & RE at $t=0.2\,\mathrm{s}$ & RE at $t=0.4\,\mathrm{s}$ \\
        \midrule
        Score-D  & 9.889e-03 & 1.932e-02 \\
        Score-I  & 9.076e-03 & 1.722e-02 \\
        FNO      & 9.987e-03 & 1.942e-02 \\
        State-DM & 1.066e-02 & 2.043e-02 \\
        \bottomrule
    \end{tabular}
\end{table}

We consider two out-of-distribution regimes. The forcing shift increases the deterministic forcing mode from $6$ to $10$, moving the energy input toward smaller spatial scales while retaining $\nu=5\times10^{-4}$. The viscosity shift increases $\nu$ from $5\times10^{-4}$ to $10^{-3}$ while retaining forcing mode $6$. Each change is applied at the start of the forecast from the same training-regime initial conditions. All remaining physical parameters, including the stochastic forcing, remain fixed. The updated controls enter the hybrid solvers through the governing equations and the state predictors through their physical inputs, with all learned parameters held fixed.

\begin{figure}[!htb]
    \centering
    \includegraphics[width=\textwidth]{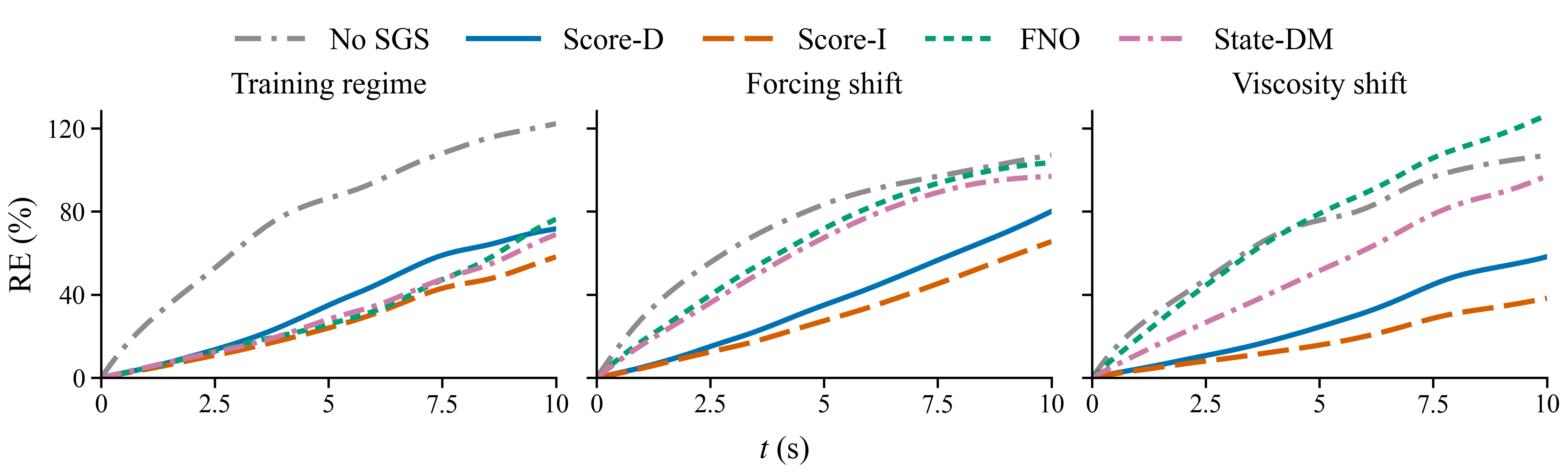}
    \caption{State prediction error growth in the training regime and under forcing and viscosity shifts. The hybrid models, Score-D and Score-I, are compared with the purely data-driven FNO and State-DM predictors and the unclosed coarse solver (No SGS). Curves show relative errors averaged over test cases, using ensemble-mean predictions for stochastic models.}
    \label{fig:regime_error_growth}
\end{figure}

Each model is evaluated over $10\,\mathrm{s}$ from $4$ held-out initial conditions, using $512$ members for stochastic predictions. We also include the unclosed coarse solver, denoted No SGS, as a baseline for assessing the effect of omitting the closure. Figure~\ref{fig:regime_error_growth} shows comparable error growth among the learned models in the training regime, followed by a clearer separation under the two parameter shifts. Under both regime shifts, both hybrid models maintain lower prediction errors than the purely data-driven models. At $10\,\mathrm{s}$, Score-I achieves RE of $0.655$ under the forcing shift and $0.383$ under the viscosity shift, compared with $1.035$ and $1.264$ for FNO and $0.969$ and $0.971$ for State-DM. Table~\ref{tab:regime_state_errors} summarizes the final prediction errors.

\begin{table}[!htb]
    \centering
    \caption{Relative state prediction errors against FDNS at $t=10\,\mathrm{s}$ in the training regime and under forcing and viscosity shifts. Errors are averaged over test cases, using ensemble-mean predictions for stochastic models. Bold values indicate the lowest error in each regime.}
    \label{tab:regime_state_errors}
    \setlength{\tabcolsep}{12pt}
    \renewcommand{\arraystretch}{1.15}
    \begin{tabular}{@{}lccc@{}}
        \toprule
        Model & Training regime & Forcing shift & Viscosity shift \\
        \midrule
        No SGS   & 1.222e+00 & 1.071e+00 & 1.070e+00 \\
        Score-D  & 7.165e-01 & 8.007e-01 & 5.823e-01 \\
        Score-I  & \textbf{5.816e-01} & \textbf{6.554e-01} & \textbf{3.826e-01} \\
        FNO      & 7.632e-01 & 1.035e+00 & 1.264e+00 \\
        State-DM & 6.888e-01 & 9.691e-01 & 9.707e-01 \\
        \bottomrule
    \end{tabular}
\end{table}

\begin{figure}[!htb]
    \centering
    \includegraphics[width=\textwidth]{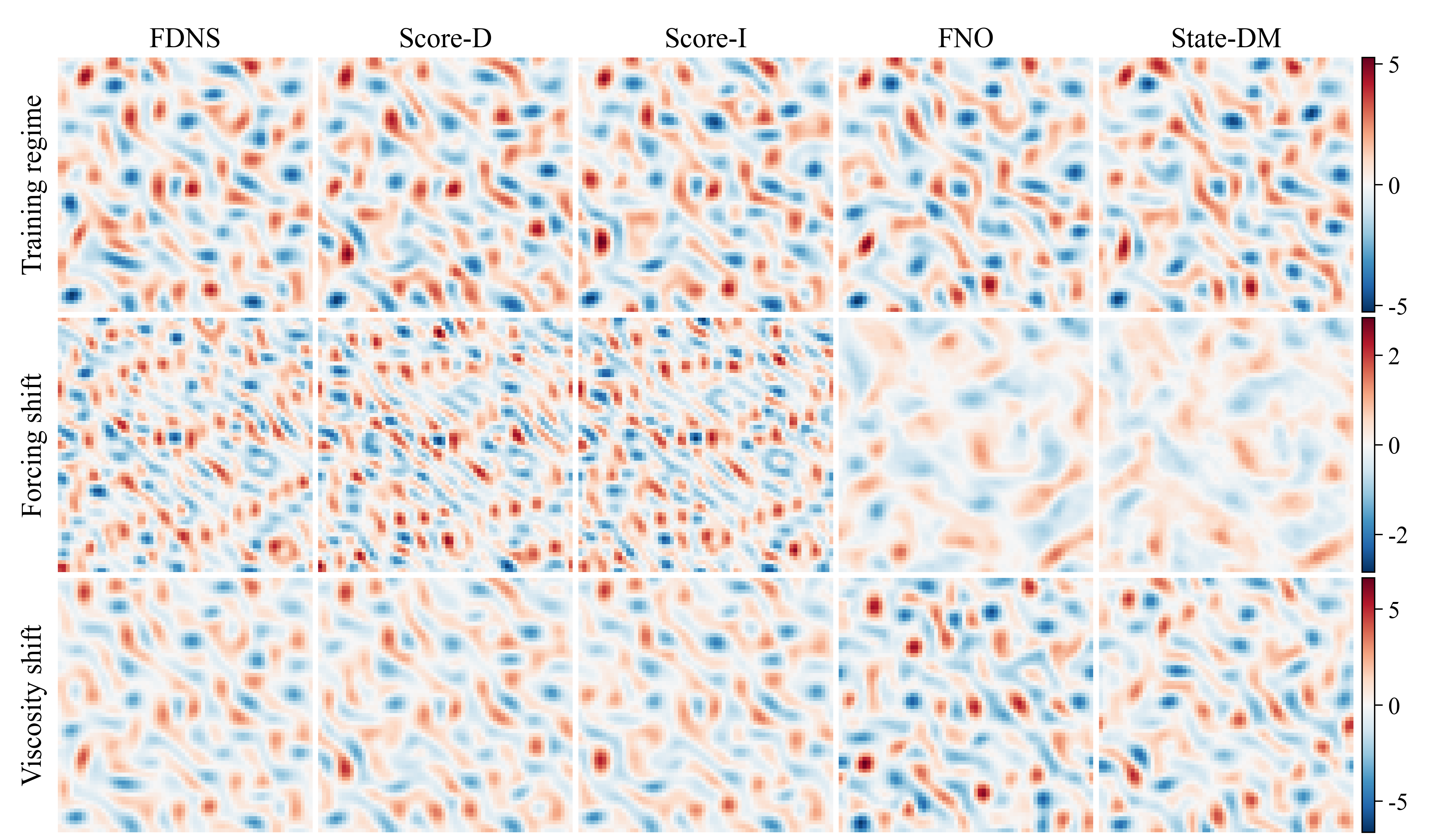}
    \caption{Resolved vorticity fields at $t=10\,\mathrm{s}$. Rows correspond to the training regime, forcing shift, and viscosity shift, from top to bottom. Columns compare FDNS with Score-D, Score-I, FNO, and State-DM.}
    \label{fig:regime_final_fields}
\end{figure}

The corresponding vorticity fields in Fig.~\ref{fig:regime_final_fields} reveal different physical consequences of the regime changes. Under the forcing shift, FDNS develops finer spatial structures that both hybrid models reproduce, while FNO and State-DM produce substantially smoother fields. Under the viscosity shift, the state predictors exhibit excessive vorticity fluctuations, whereas the hybrid predictions remain closer to the reference amplitudes. Thus, the differences in state prediction accuracy are accompanied by discrepancies in the spatial structure and intensity of the simulated flow.

These differences also appear in the resolved kinetic energy and enstrophy. For each test case, we measure their relative deviations from the FDNS ensemble statistics as
\begin{equation}\label{eq:physical_statistic_deviation}
    \delta_{\mathcal{Q}}(t)=\,\frac{\mu_{\mathcal{Q}}(t)-\mu_{\mathcal{Q}}^{\mathrm{ref}}(t)}{\mu_{\mathcal{Q}}^{\mathrm{ref}}(t)}, \qquad \mathcal{Q}\in\{\mathcal{E},\mathcal{Z}\},
\end{equation}
where $\mu_{\mathcal{Q}}$ and $\mu_{\mathcal{Q}}^{\mathrm{ref}}$ are ensemble means of statistics evaluated for individual model and FDNS realizations, respectively.

\begin{figure}[!htb]
    \centering
    \includegraphics[width=\textwidth]{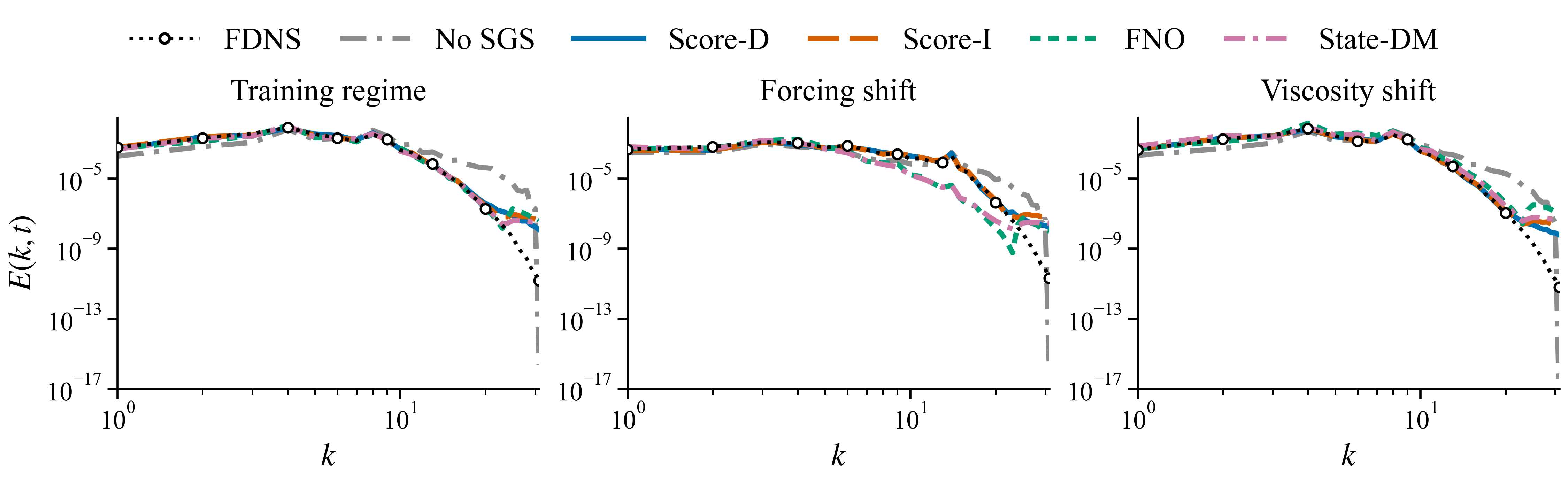}
    \caption{Resolved kinetic energy spectra $E(k)$ at $t=10\,\mathrm{s}$ in the training regime and under forcing and viscosity shifts, from left to right. Curves compare the model predictions with FDNS, with spectra averaged over realizations and test cases.}
    \label{fig:regime_energy_spectra}
\end{figure}

\begin{table}[!htb]
    \centering
    \caption{Relative deviations of resolved kinetic energy $\mathcal{E}$ and enstrophy $\mathcal{Z}$ from FDNS at $t=10\,\mathrm{s}$ under forcing and viscosity shifts. Deviations are evaluated for each test case and then averaged. Bold values indicate the smallest absolute deviation in each column.}
    \label{tab:regime_physical_deviations}
    \setlength{\tabcolsep}{12pt}
    \renewcommand{\arraystretch}{1.15}
    \begin{tabular}{@{}lcccc@{}}
        \toprule
        & \multicolumn{2}{c}{Forcing shift}
        & \multicolumn{2}{c}{Viscosity shift} \\
        \cmidrule(lr){2-3}\cmidrule(lr){4-5}
        Model & $\mathcal{E}$ & $\mathcal{Z}$
              & $\mathcal{E}$ & $\mathcal{Z}$ \\
        \midrule
        Score-D
        & -4.520e-02 & \textbf{+9.900e-02}
        & +2.550e-02 & +7.750e-02 \\
        Score-I
        & \textbf{-4.100e-02} & +1.163e-01
        & \textbf{+2.400e-02} & \textbf{+3.740e-02} \\
        FNO
        & -1.217e-01 & -6.085e-01
        & +6.394e-01 & +7.659e-01 \\
        State-DM
        & -2.401e-01 & -6.823e-01
        & +2.832e-01 & +3.502e-01 \\
        \bottomrule
    \end{tabular}
\end{table}
Table~\ref{tab:regime_physical_deviations} shows that both hybrid models retain substantially closer energy and enstrophy statistics under the two regime shifts. Under the forcing shift, the pronounced enstrophy deficits of FNO and State-DM are consistent with their loss of fine-scale structures. Under the viscosity shift, both state predictors overestimate energy and enstrophy, while the hybrid models capture the increased dissipation through the governing equations without retraining. Figure~\ref{fig:regime_energy_spectra} shows how these differences appear across spatial scales. Both hybrid models reproduce the FDNS energy distribution more closely over the energy-containing wavenumbers, particularly under the forcing shift.

In conclusion, starting from comparable short-window accuracy, both hybrid models give lower state errors and more accurate physical statistics under forcing and viscosity shifts than purely data-driven models. Score-I retains these benefits without closure labels and, within the matched comparison, achieves lower long-horizon state errors even than Score-D. This behavior is consistent with its trajectory-based objective, which evaluates closure samples through their effects on the resolved evolution. We next examine how observations acquired during simulation can further improve these predictions.

\subsection{Online data assimilation at the model-form uncertainty and prediction levels}\label{ssec:data_assimilation}

Sampled closures drive the resolved evolution, and the predicted state supplies the conditions for subsequent closure sampling. Errors in the closure model and predicted state can therefore reinforce each other over time. Resolved-state observations can constrain this feedback at two levels. At the model-form uncertainty level, observations refine the closure distribution indirectly through the resolved solver, favoring closures whose simulated evolution agrees with the measurements. At the prediction level, state EnKF corrects errors remaining in the resolved field. We first compare posterior-score DA and closure EnKF during sequential assimilation of sparse vorticity observations, examining their effects on state prediction accuracy and closure statistics. We then assess how state EnKF complements closure conditioning in improving prediction accuracy and uncertainty calibration. Finally, we use a designed case with nonlinear speed observations to demonstrate the advantage of posterior-score DA over closure EnKF in capturing non-Gaussian posterior distributions.

\subsubsection{Sequential assimilation of sparse state observations}\label{sssec:sequential_da}

We begin with sequential assimilation of sparse, linear vorticity observations, examining how repeated closure updates affect closure statistics and state prediction accuracy over time. Simulations span $20\,\mathrm{s}$ in the training regime and two forcing-transfer regimes, with forcing modes $6$, $4$, and $10$, respectively. Within each regime, we evaluate all methods from $4$ held-out initial conditions. For each initial condition, all methods start from the same resolved field and assimilate the same observation sequence using ensembles of $512$ members. Resolved vorticity is observed at $16\times16$ uniformly spaced points every $0.05\,\mathrm{s}$, with independent Gaussian noise whose standard deviation is $1\%$ of the initial root mean square vorticity at the sensors. The trained Score-I model remains fixed throughout.

\begin{figure}[!t]
    \centering
    \includegraphics[width=\textwidth]{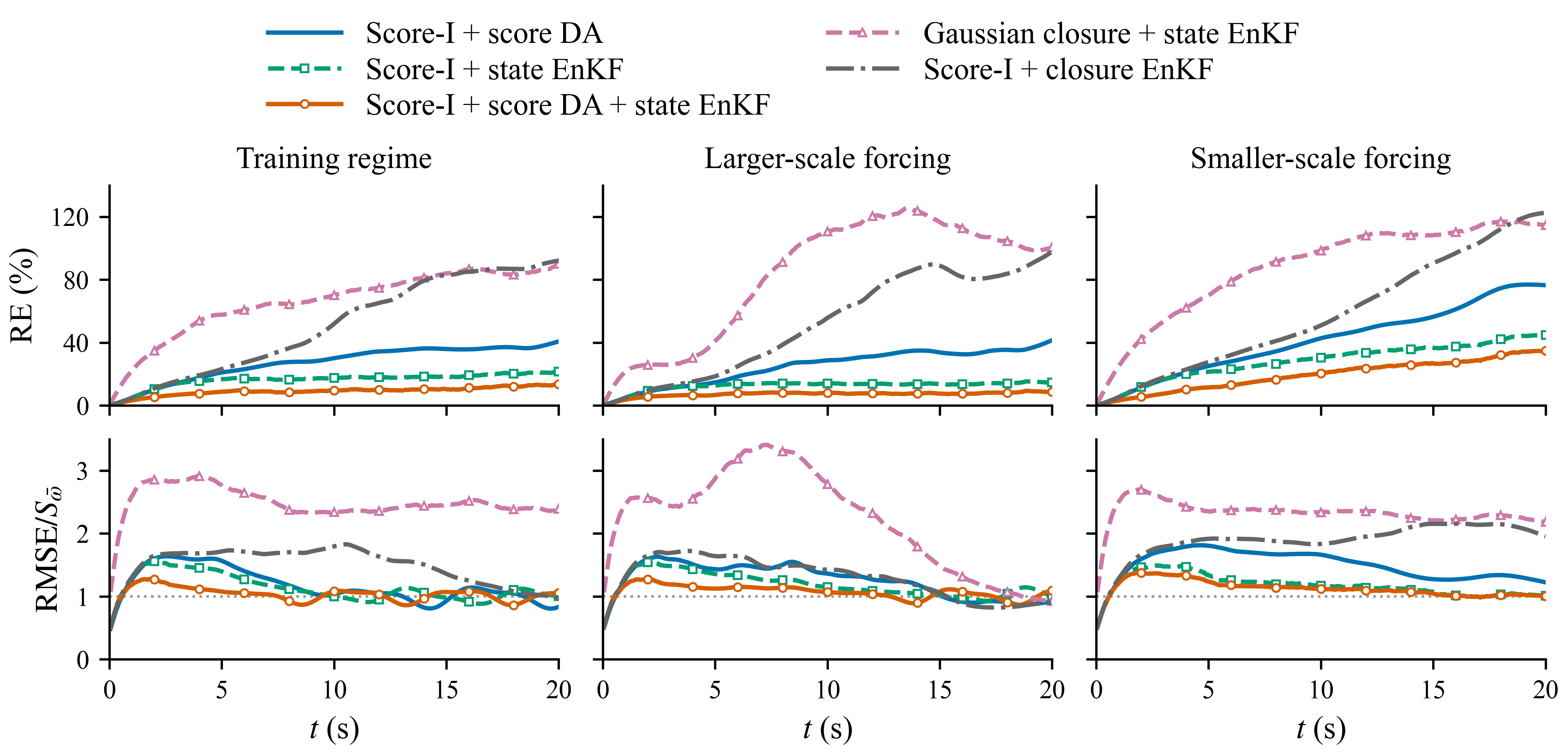}
    \caption{Sequential data assimilation in the training, larger-scale forcing, and smaller-scale forcing regimes, from left to right. Upper panels show relative errors of ensemble-mean state predictions against FDNS; lower panels show the ratio of RMSE to ensemble spread $S_{\bar{\omega}}$. The dotted horizontal line marks equal RMSE and spread. Sparse vorticity observations are assimilated every $0.05\,\mathrm{s}$.}
    \label{fig:da_error_growth}
\end{figure}

We compare posterior-score DA with the closure EnKF baseline, using Score-I as the closure model in both cases. To assess the contribution of state correction, we also consider state EnKF on its own and in combination with posterior-score DA. For the combined method, we set $\alpha_{\mathrm{DA}}=0.5$ in Eq.~\eqref{eq:method_likelihood_allocation}. An additional baseline replaces Score-I with a Gaussian random-field closure fitted to the mean and Fourier power spectrum of the training closures, while retaining state EnKF. Figure~\ref{fig:da_error_growth} shows that posterior-score DA alone consistently limits error growth more effectively than closure EnKF. While their predictions remain relatively close initially, the difference increases as simulation and assimilation continue. At $t=20\,\mathrm{s}$, posterior-score DA reduces RE relative to closure EnKF by approximately $56\%$, $58\%$, and $38\%$ in the training, larger-scale forcing, and smaller-scale forcing regimes, respectively. Table~\ref{tab:da_final_errors} summarizes the final state errors for all configurations.

\begin{table}[!htb]
    \centering
    \caption{Relative errors of ensemble-mean state predictions against FDNS at $t=20\,\mathrm{s}$ for different DA strategies. Columns compare the training, larger-scale forcing, and smaller-scale forcing regimes. Score-I supplies the closure model except for the Gaussian-closure baseline. Bold values indicate the lowest error in each regime.}
    \label{tab:da_final_errors}
    \setlength{\tabcolsep}{8pt}
    \renewcommand{\arraystretch}{1.15}
    \begin{tabularx}{\linewidth}{@{}>{\raggedright\arraybackslash}Xccc@{}}
        \toprule
        Method
        & \shortstack{Training\\regime}
        & \shortstack{Larger-scale\\forcing}
        & \shortstack{Smaller-scale\\forcing} \\
        \midrule
        Score-I only
        & 8.628e-01 & 9.528e-01 & 9.315e-01 \\
        Score DA
        & 4.072e-01 & 4.147e-01 & 7.648e-01 \\
        Closure EnKF
        & 9.214e-01 & 9.789e-01 & 1.225e+00 \\
        State EnKF
        & 2.156e-01 & 1.468e-01 & 4.491e-01 \\
        Score DA + state EnKF
        & \textbf{1.356e-01} & \textbf{8.683e-02} & \textbf{3.491e-01} \\
        Gaussian closure + state EnKF
        & 9.016e-01 & 1.007e+00 & 1.147e+00 \\
        \bottomrule
    \end{tabularx}
\end{table}

\begin{figure}[!t]
    \centering
    \includegraphics[width=\textwidth]{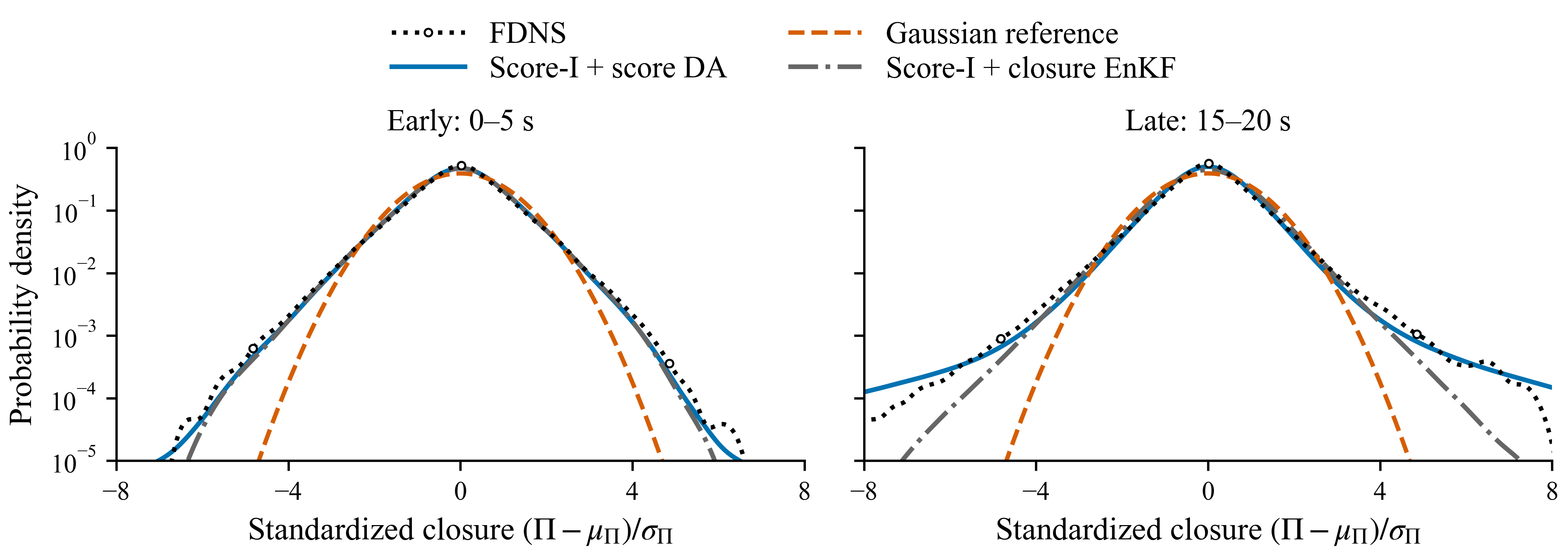}
    \caption{Standardized closure probability densities during the early ($0$--$5\,\mathrm{s}$) and late ($15$--$20\,\mathrm{s}$) stages of training-regime assimilation. Panels compare FDNS closures with samples from Score-I using posterior-score DA or closure EnKF. The standard Gaussian density is included for comparison with the non-Gaussian closure distributions.}
    \label{fig:training_closure_standardized_pdf}
\end{figure}

The growing difference in state prediction accuracy is accompanied by changes in the closure distributions. Figure~\ref{fig:training_closure_standardized_pdf} compares posterior closure samples from the early and late stages of the training-regime simulations with reference closures evaluated from the underlying DNS fields using Eq.~\eqref{eq:closure_definition}. The reference values are pooled over spatial points and times along the FDNS trajectory, while the generated values are additionally pooled over ensemble members. Each distribution is then standardized by its own mean and standard deviation to compare their shapes. The standard deviations reported in Table~\ref{tab:da_closure_statistics} are computed from all samples before standardization, whereas excess kurtosis uses samples within the common standardized interval where the estimated FDNS density is at least $10^{-5}$.

The two methods produce similar standardized closure distributions initially. At later times, posterior-score DA better reproduces the heavier tails of the reference closures, whereas closure EnKF underrepresents them. Table~\ref{tab:da_closure_statistics} quantifies these differences using closure standard deviations and excess kurtosis. The late-time excess kurtosis is $5.23$ for posterior-score DA and $1.52$ for closure EnKF, compared with $4.25$ for FDNS. The corresponding standard deviations are $0.620$ and $0.974$, respectively, compared with $0.453$ for FDNS. These results show that posterior-score DA better preserves the non-Gaussian statistics of the reference closures while achieving lower state prediction errors.

\begin{table}[!htb]
    \centering
    \caption{Closure standard deviations $\sigma_{\Pi}$ and excess kurtosis for FDNS and Score-I with posterior-score DA or closure EnKF. Statistics are compared over the early ($0$--$5\,\mathrm{s}$) and late ($15$--$20\,\mathrm{s}$) stages of training-regime assimilation.}
    \label{tab:da_closure_statistics}
    \begin{tabular}{lcccc}
        \toprule
        & \multicolumn{2}{c}{Early: $0$--$5\,\mathrm{s}$}
        & \multicolumn{2}{c}{Late: $15$--$20\,\mathrm{s}$} \\
        \cmidrule(lr){2-3}\cmidrule(lr){4-5}
        Method & $\sigma_{\Pi}$ & Excess kurtosis
               & $\sigma_{\Pi}$ & Excess kurtosis \\
        \midrule
        FDNS
        & 0.477 & 2.00 & 0.453 & 4.25 \\
        Score-I + score DA
        & 0.510 & 1.52 & 0.620 & 5.23 \\
        Score-I + closure EnKF
        & 0.505 & 1.48 & 0.974 & 1.52 \\
        \bottomrule
    \end{tabular}
\end{table}

\subsubsection{Complementary state correction and uncertainty calibration}\label{sssec:state_correction}

Assimilation at the model-form uncertainty level uses observations to refine the closures driving the resolved flow, but errors can remain in the predicted state. State EnKF complements closure conditioning by correcting these errors directly at the prediction level. Figure~\ref{fig:da_error_growth} and Table~\ref{tab:da_final_errors} show that both approaches effectively control error growth, with state EnKF achieving lower prediction errors. Combining the two substantially improves upon closure conditioning alone and reduces the final RE relative to standalone state EnKF by approximately $37\%$, $41\%$, and $22\%$ in the training, larger-scale forcing, and smaller-scale forcing regimes, respectively. These gains show that refining the closure distribution remains beneficial even when the resolved state is directly corrected. The much larger errors obtained with the Gaussian closure and state EnKF show that matching the training closure mean and Fourier power spectrum alone is insufficient for accurate state predictions, even when state observations are assimilated.

\begin{figure}[!t]
    \centering
    \includegraphics[width=\textwidth]{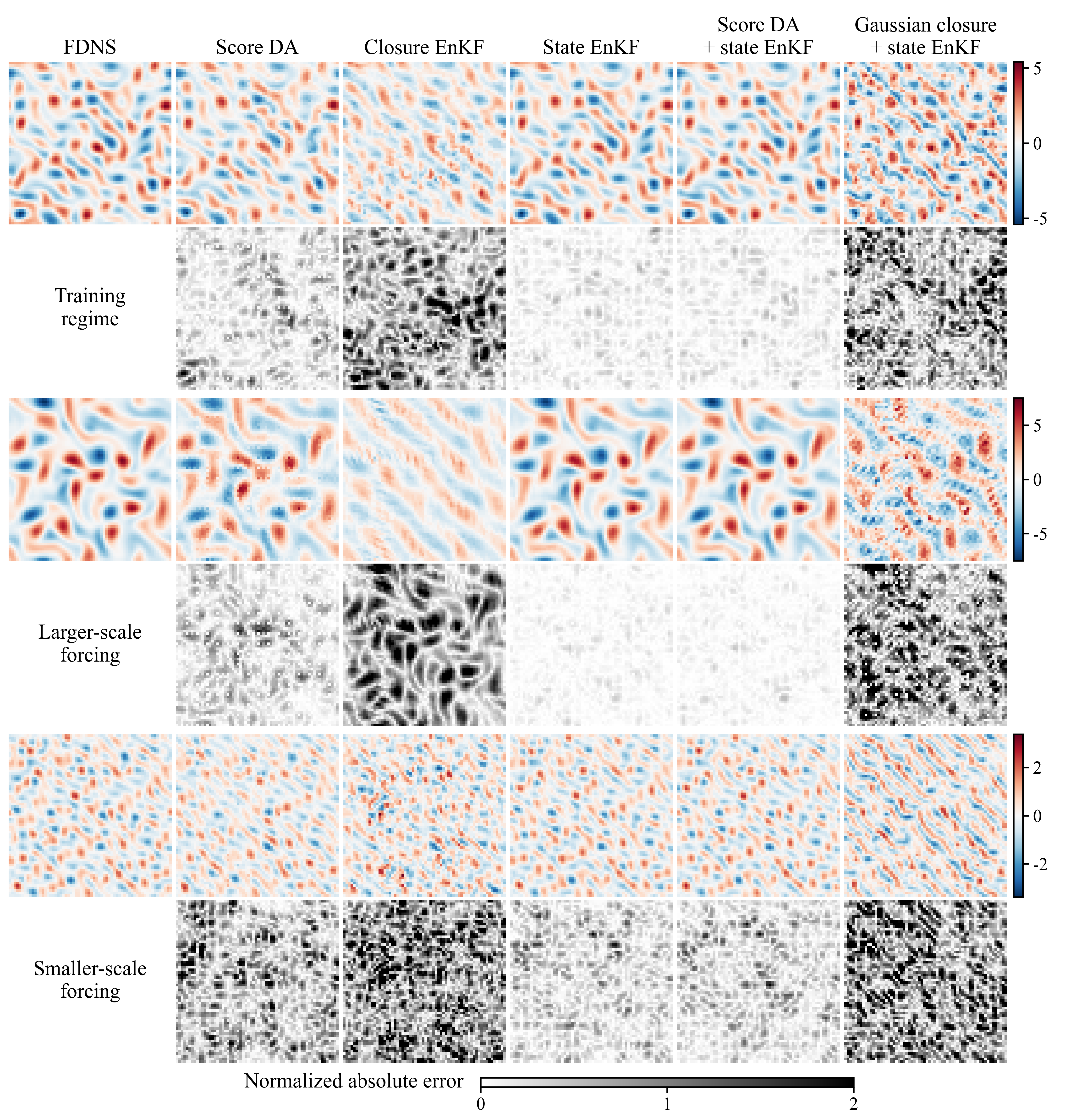}
    \caption{Resolved vorticity predictions and their errors relative to FDNS at $t=20\,\mathrm{s}$ for different DA strategies. Each pair of rows shows the vorticity fields followed by their corresponding errors. Row pairs represent the training, larger-scale forcing, and smaller-scale forcing regimes, from top to bottom. Columns compare FDNS with the five assimilation configurations, using ensemble-mean fields for the predictions.}
    \label{fig:da_final_fields}
\end{figure}

These differences are also visible in the final vorticity fields in Fig.~\ref{fig:da_final_fields}. Posterior-score DA preserves the reference flow structures more accurately than closure EnKF, and state correction further reduces their amplitude and spatial discrepancies. Agreement is particularly close in the training and larger-scale forcing regimes. More noticeable discrepancies remain under smaller-scale forcing, where finer spatial structures are harder to reconstruct from sparse observations. In this more challenging regime, however, posterior-score DA still captures the main flow variations, and complementary state correction further reduces prediction errors, as shown in Table~\ref{tab:da_final_errors}.

Data assimilation also accelerates uncertainty calibration, as measured by the balance between ensemble-mean error and spread. Since all members start from the same resolved field, ensemble spread develops as the simulations evolve. In the training regime, posterior-score DA approaches an RMSE-to-spread ratio of approximately $1$ around $t=10\,\mathrm{s}$, compared with $t=16\,\mathrm{s}$ for the unassimilated Score-I forecasts in Fig.~\ref{fig:training_rollout_diagnostics}. Combining closure conditioning with state correction brings this balance forward to approximately $t=6\,\mathrm{s}$ and maintains it through the later stages of the simulation. Similar behavior occurs in the larger-scale forcing regime, as shown in the lower panels of Fig.~\ref{fig:da_error_growth}. Both standalone state EnKF and the combined method fluctuate around $1$ during the later stages in these two regimes, while the combined method achieves lower state prediction errors. Under smaller-scale forcing, posterior-score DA alone remains underdispersed, whereas adding state EnKF brings the RMSE-to-spread ratio close to $1$ during the later stages.

\begin{figure}[!t]
    \centering
    \includegraphics[width=\textwidth]{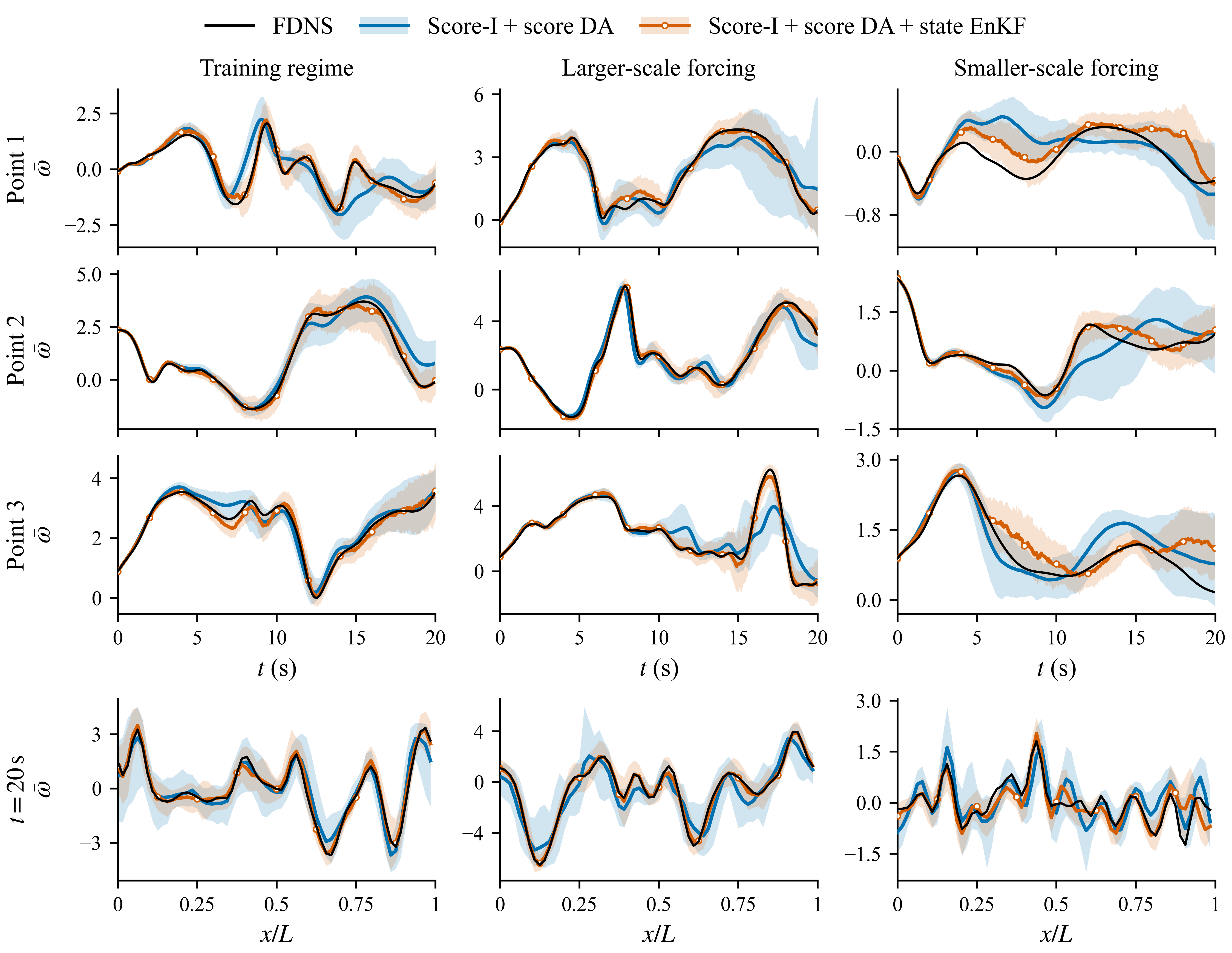}
    \caption{Comparison of posterior-score DA and its combination with state EnKF at unobserved locations. Columns correspond to the training, larger-scale forcing, and smaller-scale forcing regimes. The first three rows show vorticity time series at three fixed unobserved points; the bottom row shows spatial profiles along an unobserved grid row at $t=20\,\mathrm{s}$. Colored curves show ensemble means, shaded bands indicate central $95\%$ ensemble intervals, and black curves show FDNS.}
    \label{fig:da_uq_combined}
\end{figure}

The benefit of state correction also extends to locations where observations are unavailable. Figure~\ref{fig:da_uq_combined} compares vorticity time series at three unobserved points and final spatial profiles along an unobserved grid row. In the training and larger-scale forcing regimes, the combined method follows the reference evolution more closely while generally maintaining narrower ensemble intervals than posterior-score DA alone. The spatial profiles show corresponding improvements in the predicted peaks and troughs. Under smaller-scale forcing, the combined method still generally follows FDNS more closely and provides better-calibrated uncertainty than posterior-score DA alone, although noticeable discrepancies remain in some fluctuations.

\subsubsection{Closure conditioning with nonlinear observations}\label{sssec:nonlinear_closure_conditioning}

The preceding experiments use linear observations of resolved vorticity, a standard setting for EnKF-based state estimation \citep{evensen2003ensemble}. These measurements constrain closures through the intervening nonlinear flow evolution. We next compare posterior-score DA and closure EnKF using a common prior supplied by Score-I and scalar speed observations, motivated by single hot-wire measurements that cannot distinguish opposite flow directions \citep{bruun1995hot}.

Starting from a common resolved vorticity $\overline{\omega}_t$, each sampled closure $\Pi$ advances the flow from $t$ to $t+\Delta t$. We then recover the forecast velocity through the streamfunction:
\begin{equation}\label{eq:speed_observation}
    \begin{aligned}
        \overline{\omega}_{t+\Delta t}
        &= \Phi_{t,t+\Delta t}(\overline{\omega}_t,\Pi),\\
        -\nabla^2\psi_{t+\Delta t}
        &= \overline{\omega}_{t+\Delta t},
        \qquad
        \overline{\mathbf{u}}_{t+\Delta t}
        =(\partial_y\psi_{t+\Delta t},-\partial_x\psi_{t+\Delta t}).
    \end{aligned}
\end{equation}
Here, $\Phi_{t,t+\Delta t}$ advances Eq.~\eqref{eq:resolved_dynamics}, and the streamfunction is obtained by solving the periodic Poisson equation. For the ease of presentation, we further denote the observation operator as $h$ which evaluates the speed after this vorticity evolution and velocity reconstruction: $h(\Pi)=\|\overline{\mathbf{u}}_{t+\Delta t}(\mathbf{x}_s); \Pi\|_2$. The measured speed is modeled as $y=h(\Pi)+\epsilon$, where $\epsilon\sim\mathcal{N}(0,\sigma_{\mathrm{obs}}^2)$.The probe $\mathbf{x}_s$ is chosen to maximize velocity variance among locations with nearly zero mean predicted velocity. We set $\sigma_{\mathrm{obs}}$ to $10\%$ of the prior root mean square speed at the probe. The 10th and 90th percentiles of the resulting prior predictive measurement distribution then define the low- and high-speed observations used in the comparison.

The speed observation is nonlinear in the velocity components and constrains their magnitude while leaving direction unresolved. At the selected probe, this ambiguity favors ring-shaped joint distributions of the velocity components, giving rise to the non-Gaussian posterior structure examined here. The posterior shape and spread can therefore change with the measured speed. For a fixed prior ensemble and observation model, however, changing the observed value only translates the EnKF analysis distribution, leaving its centered shape and spread unchanged.

Conditioned on the same resolved state $\overline{\omega}_t$, we draw $M=2048$ closure samples from the Score-I prior. Closure EnKF updates these samples using Eq.~\eqref{eq:method_enkf_analysis}, while posterior-score DA uses the single-closure form of Eq.~\eqref{eq:method_posterior_score}. We also construct a Monte Carlo reference for the Bayesian posterior from the same prior, with normalized likelihood weights:
\begin{equation}\label{eq:mc_posterior_reference}
    \beta_m(y)\propto
    \exp\!\left[-\frac{(y-h(\Pi^{(m)}))^2}{2\sigma_{\mathrm{obs}}^2}\right],
    \qquad \sum_{m=1}^{M}\beta_m(y)=1,
\end{equation}
where $\Pi^{(m)}$ denotes a prior sample from the ensemble.

\begin{figure}[!htb]
    \centering
    \includegraphics[width=\textwidth]{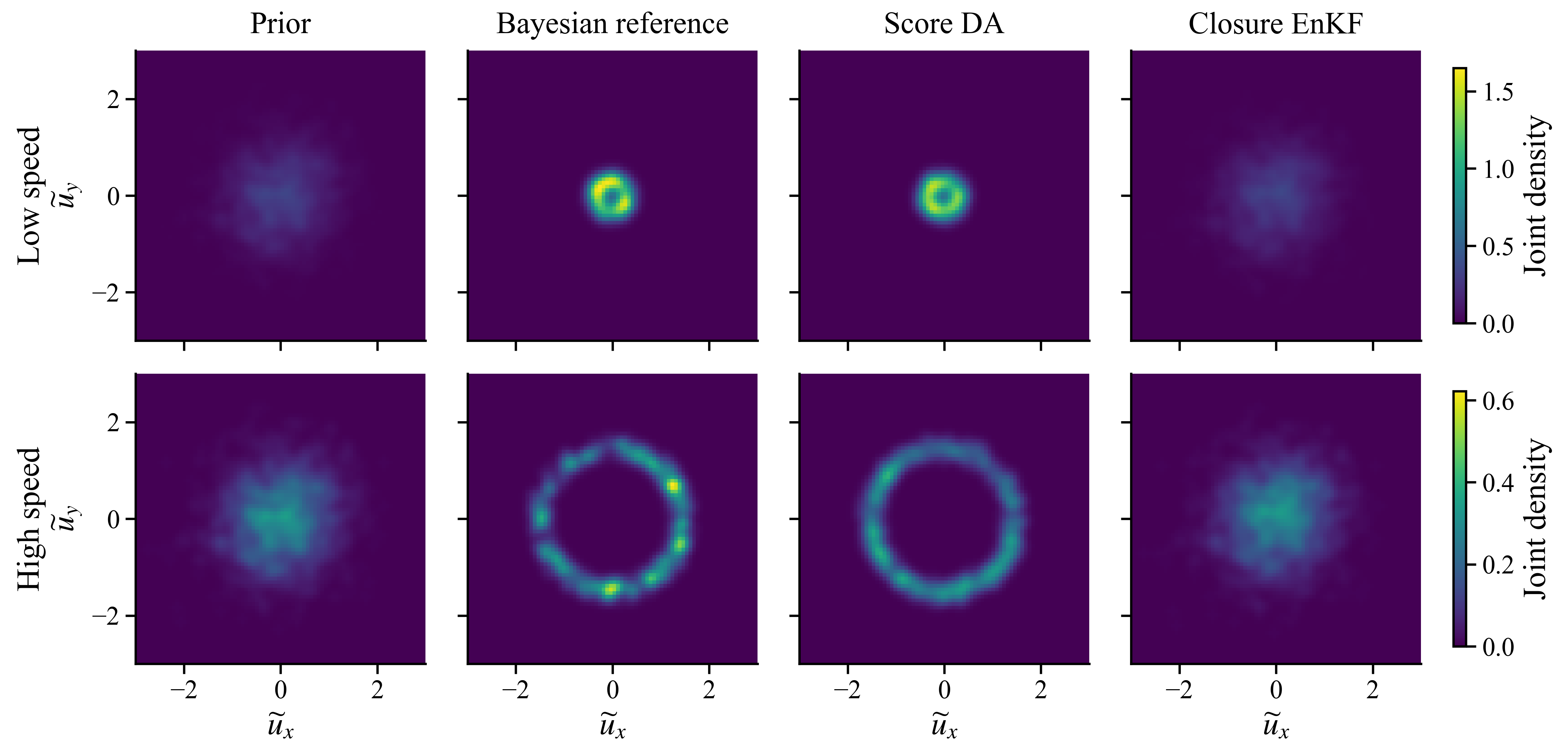}
    \caption{Joint probability densities of the normalized forecast velocity components $\widetilde{u}_x$ and $\widetilde{u}_y$ at the speed probe. Columns compare the prior, Bayesian reference, posterior-score DA, and closure EnKF. The upper and lower rows correspond to low- and high-speed observations, respectively. Both observation cases use the same Score-I prior, repeated in the first column for comparison.}
    \label{fig:speed_posterior_structure}
\end{figure}

Figure~\ref{fig:speed_posterior_structure} compares the forecast velocity distributions obtained with posterior-score DA and closure EnKF, which use the same speed observations to update closures from a common prior. Score DA reproduces the reference ring structure, whereas closure EnKF retains the prior's concentration near zero velocity. This difference reflects the difficulty of capturing the directional ambiguity of speed observations through linear cross-covariances: opposite velocity responses can produce similar speeds, causing positive and negative covariance contributions to largely cancel. In the $2048$-member prior ensemble, the cross-covariance between the closure at the probe and predicted speed is $-9.81\times10^{-9}$, consistent with the very weak EnKF correction. The total velocity variance, defined as $\operatorname{Var}(\widetilde{u}_x)+\operatorname{Var}(\widetilde{u}_y)$, further quantifies the change in posterior spread. From low to high speed, it increases from $0.127$ to $2.118$ in the Bayesian reference and from $0.132$ to $2.218$ with Score DA, while closure EnKF gives $0.998$ in both cases, nearly unchanged from the prior value of $1$. 

\begin{figure}[!htb]
    \centering
    \includegraphics[width=\textwidth]{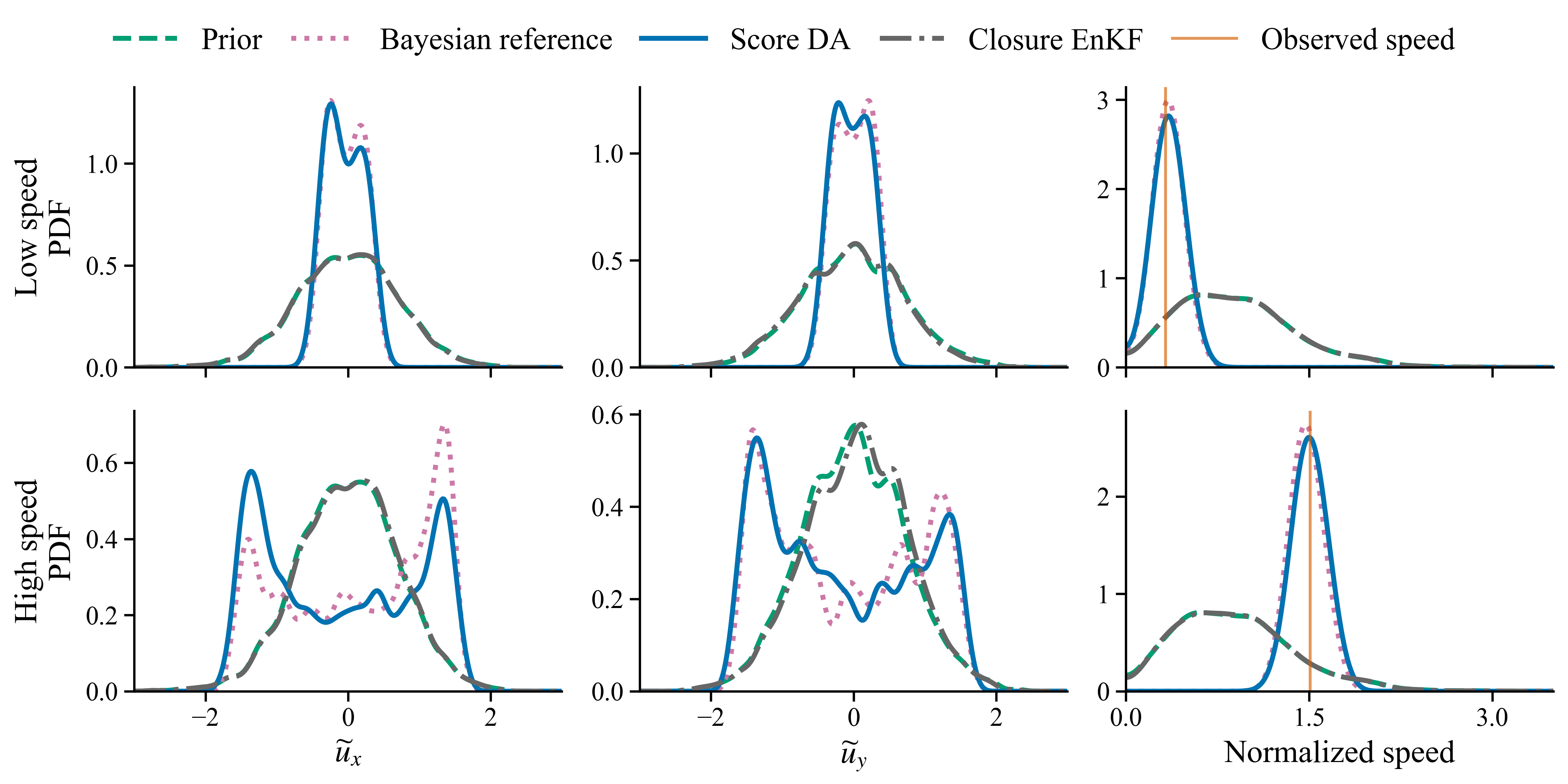}
    \caption{Prior and posterior marginal densities under nonlinear speed observations. Columns show the forecast velocity components $\widetilde{u}_x$ and $\widetilde{u}_y$ and the normalized speed at the probe, from left to right. The upper and lower rows correspond to low- and high-speed observations, respectively. Curves compare the prior, Bayesian reference, posterior-score DA, and closure EnKF; vertical lines mark the observed speeds.}
    \label{fig:speed_posterior_pdfs}
\end{figure}

Figure~\ref{fig:speed_posterior_pdfs} further compares the marginal distributions of the velocity components and the predicted speed. Score DA captures the concentrated velocity-component distributions at low speed and their two prominent peaks at high speed, while its forecast-speed densities closely follow the Bayesian reference near the observations. The closure EnKF curves remain close to the prior for both observations, indicating that the covariance-based update responds only weakly to the measured speed. Together, the joint and marginal comparisons show that posterior-score DA captures non-Gaussian posterior structure, including bimodal velocity marginals and observation-dependent spread, more accurately than closure EnKF.

\section{Conclusions}\label{sec:conclusions}

Simulating multiscale dynamical systems at an affordable cost often requires closure models to account for the effects of unresolved scales. Compared to deterministic closures with uncertain parameters, stochastic closures provide a more fundamental and systematic way of representing the variability of these effects. However, stochastic closures require the learning of conditional distributions, for which non-Gaussian statistics are often non-negligible and the reference closure data are usually unavailable in many real-world science and engineering applications. In this work, we develop a conditional score-based framework for learning stochastic closures and refining their predictions through online data assimilation. The stochastic closure model is first learned offline through indirect data of numerically resolved system states. The learned model then supports online assimilation via a progressive correction approach at two complementary levels, where posterior-score DA first uses resolved-state observations to correct closures at the model-form uncertainty level without being restricted by a Gaussian assumption, and standard EnKF further leverages the ensemble of closures and updates the prediction at the system-state level. Numerical experiments on two-dimensional turbulence demonstrate that offline indirect learning captures dominant closure structures and their contributions to resolved energy and enstrophy, producing substantial improvements in coarse-grid predictions. Comparisons with purely data-driven models further demonstrate the merit of retaining known physics for generalization capability across different regimes. For online data assimilation, posterior-score DA better reproduces the non-Gaussian statistics of the reference closures and achieves lower state prediction errors than closure EnKF, which is restricted by the Gaussian assumption. Combining posterior-score DA for closure assimilation with state-level EnKF further improves prediction and uncertainty quantification of the system state. 

\section*{Acknowledgments}

X.D. and J.W. acknowledge support from the Office of Naval Research under grant No.~N00014-24-1-2391. 

\appendix
\section{Statistical justification of the trajectory energy score loss}
\label{app:training_details}

This appendix justifies the trajectory energy score loss $\mathcal{L}_{\mathrm{ES}}$ in Eq.~\eqref{eq:trajectory_energy_estimator}. We show that its finite-ensemble expression is an unbiased estimator of the energy score and that, in expectation over reference trajectories, this score is minimized when the generated and reference trajectory distributions agree.

Fix an initial resolved field and let $P_\theta$ and $P_{\mathrm{ref}}$ denote the generated and reference distributions of the future trajectories defined in Section~\ref{ssec:closure_training}, both with finite first moments. Let $Z,Z'\sim P_\theta$ and $Y,Y'\sim P_{\mathrm{ref}}$ be mutually independent trajectories. All distances use the trajectory norm $\|\cdot\|_{\mathrm{tr}}$ defined in the main text. For a fixed recorded reference trajectory $y=Z^{\mathrm{ref}}$, consider $n\geq2$ independent generated trajectories. Each of the $n$ distances to $y$ has expectation $\mathbb{E}\|Z-y\|_{\mathrm{tr}}$, and each of the $n(n-1)$ ordered pairs of different ensemble members has expected distance $\mathbb{E}\|Z-Z'\|_{\mathrm{tr}}$. Applying linearity of expectation to Eq.~\eqref{eq:trajectory_energy_estimator} therefore gives
\begin{equation}\label{eq:appendix_estimator_connection}
    \mathbb{E}_{\mathrm{sampling}}\left[\mathcal{L}_{\mathrm{ES}}\right]=\mathbb{E}\|Z-y\|_{\mathrm{tr}}-\frac12\mathbb{E}\|Z-Z'\|_{\mathrm{tr}}\equiv \mathrm{ES}(P_\theta,y),
\end{equation}
where $\mathbb{E}_{\mathrm{sampling}}$ averages over the generated ensemble, and $\mathrm{ES}(P_\theta,y)$ is the energy score for the predictive distribution $P_\theta$ and reference realization $y$. Thus, the finite-ensemble loss is unbiased for any $n\geq2$. The required independence is between ensemble members, which use independent OU sequences; temporal correlation within each trajectory is retained.

We next average over reference trajectories $Y\sim P_{\mathrm{ref}}$. The difference between the expected scores of the generated and reference distributions is
\begin{align}
    &\mathbb{E}_{Y}\!\left[\mathrm{ES}(P_\theta,Y)\right] -\mathbb{E}_{Y}\!\left[\mathrm{ES}(P_{\mathrm{ref}},Y)\right] \nonumber\\
    &\qquad= \mathbb{E}\|Z-Y\|_{\mathrm{tr}} -\frac12\mathbb{E}\|Z-Z'\|_{\mathrm{tr}} -\frac12\mathbb{E}\|Y-Y'\|_{\mathrm{tr}} \nonumber\\
    &\qquad=\frac12\operatorname{ED}(P_\theta,P_{\mathrm{ref}}), \label{eq:appendix_energy_distance}
\end{align}
where $\operatorname{ED}$ is the energy distance. For the Euclidean trajectory norm, this distance is nonnegative and equals zero if and only if $P_\theta=P_{\mathrm{ref}}$ \citep{gneiting2007strictly,szekely2013energy}. Consequently, the expected energy score is minimized by the reference trajectory distribution. The fixed coefficient $1/2$ in Eq.~\eqref{eq:trajectory_energy_estimator} gives precisely this relation. These results establish the statistical basis of the energy-score term in Eq.~\eqref{eq:indirect_training_objective}.

\section{Numerical solver and spectral diagnostics}
\label{app:solver}

\subsection{Pseudospectral discretization and time integration}

The DNS and coarse simulations use the same Fourier pseudospectral discretization on the periodic domain. Spatial derivatives are evaluated with physical wavenumbers $\mathbf{k}=(2\pi/L)(m_x,m_y)$. For $\mathbf{k}\neq0$, the streamfunction and velocity coefficients are
\begin{equation}\label{eq:appendix_spectral_velocity}
    \widehat{\psi} =\frac{\widehat{\omega}}{|\mathbf{k}|^2}, \qquad \widehat{\mathbf{u}} =(\mathrm{i}k_y,-\mathrm{i}k_x)\widehat{\psi}.
\end{equation}
The zero-mode velocity is set to zero. Advection is evaluated by transforming the velocity and vorticity gradients to physical space, multiplying them pointwise, and transforming the result back to Fourier space. The resulting advection coefficients are multiplied by a componentwise two-thirds mask to reduce aliasing \citep{orszag1971elimination,canuto2006spectral}.

For the coarse dynamics in Eq.~\eqref{eq:resolved_dynamics}, define
$\lambda_{\mathbf{k}}=\nu|\mathbf{k}|^2+\alpha$ and
$N^r=-\widehat{\overline{\mathbf{u}}\cdot\nabla\overline{\omega}}^{\,r}
-\widehat{\Pi}^{\,r}+\widehat{\overline{f}}$.
The Crank--Nicolson/Adams--Bashforth update is
\begin{equation}\label{eq:appendix_cnab2}
    \widehat{\overline{\omega}}^{\,r+1} = \frac{(1-\tfrac12\Delta t\lambda_{\mathbf{k}}) \widehat{\overline{\omega}}^{\,r} +\Delta t(\tfrac32N^r-\tfrac12N^{r-1})}{1+\tfrac12\Delta t\lambda_{\mathbf{k}}}.
\end{equation}
Here, $r$ indexes numerical integration steps, and the first update uses $N^0$ for the explicit term. The same discretization is used for indirect training and subsequent simulations. The previous explicit right-hand side is retained across closure refreshes and recorded snapshot times, so neither event restarts the multistep integrator.

\subsection{Closure power and transfer spectra}
\label{app:metrics}

Let $\mathcal{B}_b$ denote the radial wavenumber bin centered at $b\Delta k$ with width $\Delta k=2\pi/L$. Using Fourier coefficients normalized by the number of grid points, the closure power and SGS energy and enstrophy transfer spectra are defined as \citep{guan2023stable,perezhogin2023generative}
\begin{align}
    C_\Pi(b\Delta k) &= \frac{1}{\Delta k} \sum_{\mathbf{k}\in\mathcal{B}_b} |\widehat{\Pi}(\mathbf{k})|^2, \label{eq:appendix_closure_power}\\
    T_E(b\Delta k) &= -\frac{1}{\Delta k} \sum_{\mathbf{k}\in\mathcal{B}_b} \operatorname{Re} \left[ \widehat{\psi}(\mathbf{k})^* \widehat{\Pi}(\mathbf{k}) \right], \label{eq:appendix_energy_transfer}\\
    T_Z(b\Delta k) &= -\frac{1}{\Delta k} \sum_{\mathbf{k}\in\mathcal{B}_b} \operatorname{Re} \left[ \widehat{\overline{\omega}}(\mathbf{k})^* \widehat{\Pi}(\mathbf{k}) \right]. \label{eq:appendix_enstrophy_transfer}
\end{align}
Here, $\psi$ is the zero-mean streamfunction associated with the resolved vorticity, and $^*$ denotes complex conjugation. Each spectrum is computed from an individual closure realization and its corresponding resolved state before averaging across samples.

\section{Model architectures and training details}
\label{app:implementation}

Score-I uses a conditional Fourier neural operator (FNO) with $24$ retained Fourier modes in each spatial direction, width $72$, and seven operator layers \citep{li2021fourier,dong2025data}. The noisy closure and resolved state are concatenated as input channels. Diffusion time is encoded using $256$ Gaussian Fourier features followed by two fully connected layers. Each operator layer combines a spectral convolution, a pointwise linear transformation, and a projected diffusion-time embedding, followed by GELU activation. Two pointwise projection layers map the final features to a single-channel output used to construct the conditional closure score.

Score-I is trained on trajectory windows containing an initial state and two future states at $0.2$ and $0.4\,\mathrm{s}$. Each sampling epoch draws eight windows with replacement from each training trajectory and shuffles them into minibatches of four windows. Each window uses four independently generated trajectories, with a new closure sampled at every physical step of size $10^{-3}\,\mathrm{s}$. Reverse sampling uses $32$ stochastic Euler--Maruyama steps, a terminal diffusion time of $10^{-3}$, and a power-law noise grid with exponent $7$. After numerical scaling, the diffusion coefficient is $g(\tau)=25^\tau$. The OU correlation time is $T_{\mathrm{OU}}=0.05\,\mathrm{s}$. The randomized adjoint uses a selection probability of $1/8$.

Score-D shares the closure-model architecture but uses direct denoising score matching with paired state--closure data. The original Score-D is trained for $80$ epochs with effective minibatch size $64$, retaining the exponential-moving-average checkpoint with the lowest validation denoising score-matching loss. Thus, the closure comparisons in Section~\ref{ssec:indirect_validation} use the best validation checkpoints obtained over the full training budgets of Score-I and Score-D. The FNO and State-DM predictors use the same spectral width, retained modes, and number of operator layers, with additional $3\times3$ depthwise convolutions in their local branches. Both use the same resolved training trajectories as Score-I and receive the prescribed physical parameters and forcing as inputs. For the regime-transfer comparisons in Section~\ref{ssec:generalization}, Score-I remains fixed, and comparator checkpoints are selected using training-regime validation data. Their independent test REs at both $0.2$ and $0.4\,\mathrm{s}$ are within $20\%$ of Score-I (Table~\ref{tab:short_window_matching}), allowing subsequent generalization to be compared from similar short-window prediction accuracy.

\section{Data assimilation implementation details}
\label{app:da_implementation}

Posterior-score DA uses the reverse-SDE sampler and noise grid described in~\ref{app:implementation}. Sequential assimilation uses $32$ reverse-diffusion steps, with likelihood guidance applied for $\tau_k\leq0.2$, corresponding to the final $13$ steps. Each observation interval contains $L=50$ physical steps of size $\Delta t=10^{-3}\,\mathrm{s}$, with a separate closure field at each step.

The OU correlation time is $T_{\mathrm{OU}}=0.05\,\mathrm{s}$, giving correlation coefficient $\rho=\exp(-\Delta t/T_{\mathrm{OU}})$ between consecutive physical steps. Initialization and each reverse-diffusion step use separate OU noise sequences, generated using independent Gaussian random variables for each ensemble member. The first interval uses stationary OU noise; subsequent intervals condition on the preceding terminal noise, giving
\begin{equation}\label{eq:appendix_ou_guidance_covariance}
    \mathcal{K}^{(j)}_{\ell q}=
    \begin{cases}
        \rho^{|\ell-q|}, & j=1,\\
        \rho^{|\ell-q|}-\rho^{\ell+q+2}, & j>1,
    \end{cases}
    \qquad \ell,q=0,\ldots,L-1.
\end{equation}
For continuation into the next interval, the terminal noise at each guided diffusion step is updated to $\xi_{L-1,k}+\sqrt{\Delta\sigma_k^2}\sum_{q=0}^{L-1}\mathcal{K}^{(j)}_{L-1,q}g^{\mathrm{lik}}_{q,k}$. Initialization noise and unguided steps continue without this adjustment.

Each likelihood evaluation and subsequent forecast using updated closures starts from the same incoming vorticity and solver history $N^{r-1}$ in Eq.~\eqref{eq:appendix_cnab2}. Closure EnKF jointly analyzes the $L$ closure fields and terminal OU noise, retaining the analyzed noise for the next interval. State EnKF jointly analyzes the forecast vorticity and the inverse Fourier transform of the stored right-hand side. Within each analysis, all updated fields share the same memberwise observation perturbations.

Sequential EnKF analyses use observation-precision localization with a periodic squared-cosine taper of support radius eight grid cells \citep{hamill2001distance,hunt2007efficient}. State analyses additionally apply relaxation to prior spread with coefficient $0.95$ to both vorticity and solver history, preserving their analyzed means \citep{whitaker2012evaluating}. Combined assimilation uses $\alpha_{\mathrm{DA}}=0.5$ in Eq.~\eqref{eq:method_likelihood_allocation}.

For the nonlinear-observation comparison in Section~\ref{sssec:nonlinear_closure_conditioning}, we generate $1000$ posterior-score samples per observation using a finer reverse-diffusion grid, with likelihood guidance applied for $\tau_k\leq0.5$. The closure-EnKF analysis uses neither localization nor spread relaxation.

\bibliographystyle{model1-num-names}
\bibliography{references.bib} 

\end{document}